\documentclass{article}
\usepackage{iclr2026_conference,times}

\usepackage[utf8]{inputenc}
\usepackage[T1]{fontenc}
\usepackage{hyperref}
\usepackage{url}
\usepackage{booktabs}
\usepackage{amsfonts}
\usepackage{amsmath}
\usepackage{amssymb}
\usepackage{nicefrac}
\usepackage{microtype}
\usepackage{xcolor}
\usepackage{graphicx}
\usepackage{subcaption}
\usepackage{algorithm}
\usepackage{algorithmic}
\usepackage{multirow}
\usepackage{wrapfig}
\usepackage{float}
\usepackage{colortbl}
\usepackage{enumitem}
\usepackage{pifont}
\usepackage{tikz}

\newcommand{\cmark}{\ding{51}}
\newcommand{\xmark}{\ding{55}}
\definecolor{myblue}{RGB}{66,133,244}

\newcommand{\method}{Ultra Flash}

\newcommand{\eg}{\textit{e.g.}}

\newcommand{\lightningbolt}{%
  \begin{tikzpicture}[baseline=-0.05ex, scale=0.22]
    \fill (0.4,2.2) -- (1.5,2.2) -- (0.7,1.2) -- (1.3,1.2) -- (0,0) -- (0.5,1.0) -- (0,1.0) -- cycle;
  \end{tikzpicture}%
}

\definecolor{cL}{HTML}{2563EB}
\definecolor{cR}{HTML}{9333EA}
\title{%
  {\color{cL}\lightningbolt}\,{\color{cL}U}%
  {\color{cL!89!cR}l}%
  {\color{cL!78!cR}t}%
  {\color{cL!67!cR}r}%
  {\color{cL!56!cR}a}
  {\color{cL!44!cR}F}%
  {\color{cL!33!cR}l}%
  {\color{cL!22!cR}a}%
  {\color{cL!11!cR}s}%
  {\color{cR}h}%
  : Scaling Real-Time Streaming Video Generation to High Resolutions%
}

\author{
Jie Huang\textsuperscript{1\dag},
Xin Lu\textsuperscript{2\dag},
Zihao Fan\textsuperscript{2\dag},
Xiaoxiao Ma\textsuperscript{2},
Yuming Li\textsuperscript{3},
Jun-hao Zhuang\textsuperscript{1},\\
\textbf{Zeyue Xue}\textsuperscript{1},
\textbf{Siming Fu}\textsuperscript{1},
\textbf{Haoran Li}\textsuperscript{1},
\textbf{Mingchen Zhong}\textsuperscript{2},
\textbf{Guohui Zhang}\textsuperscript{2},\\
\textbf{Shichen Ma}\textsuperscript{1},
\textbf{Yijun Liu}\textsuperscript{4},
\textbf{Jiaqi Shi}\textsuperscript{2},
\textbf{Yanwen Ma}\textsuperscript{5},
\textbf{Yaofeng Su}\textsuperscript{6},
\textbf{Haoyu Wang}\textsuperscript{4},\\
\textbf{Yaowei Li}\textsuperscript{3},
\textbf{Songchun Zhang}\textsuperscript{7},
\textbf{Weiyang Jin}\textsuperscript{8},
\textbf{Yuxuan Bian}\textsuperscript{9},
\textbf{Shiyi Zhang}\textsuperscript{4},
\textbf{Haojun Xu}\textsuperscript{5},\\
\textbf{Shuai Lu}\textsuperscript{1},
\textbf{Xin Han}\textsuperscript{1},
\textbf{Wei Tang}\textsuperscript{1},
\textbf{Haoyang Huang}\textsuperscript{1},
\textbf{Nan Duan}\textsuperscript{1},
\textbf{Xueyang Fu}\textsuperscript{2,\ding{41}}\\
\textsuperscript{1}JD Explore Academy,
\textsuperscript{2}USTC,
\textsuperscript{3}PKU,
\textsuperscript{4}THU,
\textsuperscript{5}BUAA,
\textsuperscript{6}FDU,
\textsuperscript{7}HKUST,
\textsuperscript{8}HKU,
\textsuperscript{9}CUHK
}

\iclrfinalcopy

\begin{document}

\maketitle

\vspace{-2em}
\begin{figure}[H]
  \centering
  \includegraphics[width=.99\textwidth]{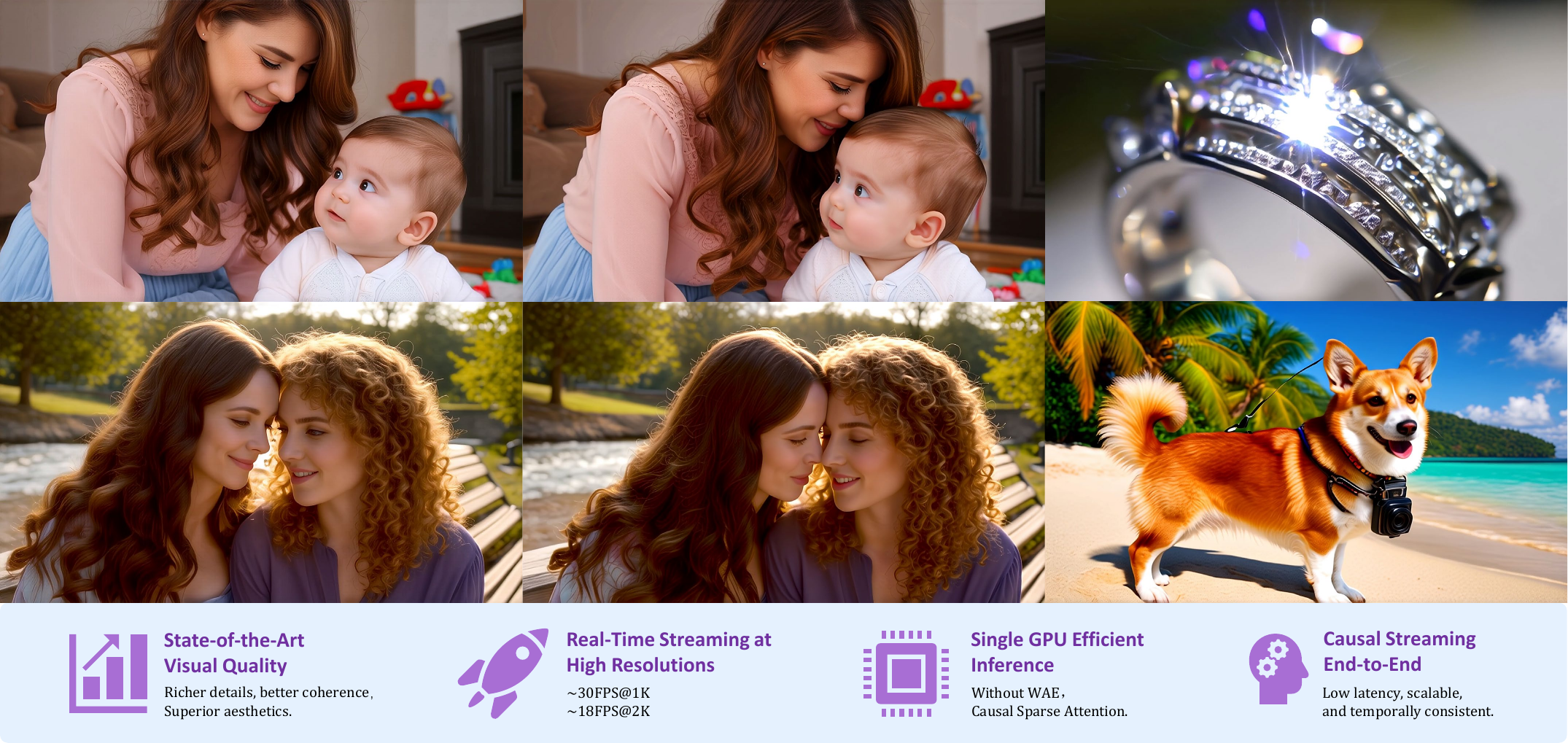}
  \label{fig:abs}
\end{figure}
\vspace{-1em}

\begin{abstract}

While recent autoregressive video diffusion models achieve remarkable streaming quality, they remain confined to low resolutions (\textit{e.g.}, $480$P), leaving efficient, scalable, real-time high-resolution video generation a fundamental open challenge. To bridge this gap, we present Ultra Flash, a cascaded streaming framework capable of real-time high-resolution video generation. Ultra Flash achieves ${\sim}30$ FPS at 1K resolution and ${\sim}18$ FPS at 2K resolution on a single GPU through three key contributions: (1) an architecture-preserving T2V-to-TV2V super-resolution training paradigm coupled with an AIGC-oriented data degradation pipeline that effectively preserves the generative capability of the base model, enabling enhanced high-resolution detail when cascaded after mainstream low-resolution generative models; (2) a causal streaming latent upsampler paired with a high-resolution decoder, which enhances spatiotemporal coherence while enabling efficient latent spatial scaling and precise high-resolution decoding with negligible computational overhead; and (3) a cascade high-resolution streaming video generation optimization scheme that first performs hybrid-reward-enhanced sparse causalization and single-step distillation of the super-resolution model, then introduces cascaded streaming self-forcing preference optimization with dynamic cache management, jointly enhancing overall coherence, improving quality, and enabling real-time high-resolution streaming video generation. Extensive experiments demonstrate that Ultra Flash reliably produces ultra-high-resolution streaming video while maintaining state-of-the-art visual quality and superior efficiency.
\paragraph{Project Page:} \url{https://xin1u.github.io/UltraFlash/}
 
\end{abstract}

\section{Introduction}

\label{sec:intro}

Video diffusion models have made extraordinary progress in generating photorealistic video from text prompts~\cite{wan2025wan,polyak2024movie,yang2024cogvideox,kong2024hunyuanvideo}. Meanwhile, interactive applications---such as real-time previewing, game asset generation, and live content creation---demand \emph{streaming} output at high resolution and low latency. Recent autoregressive adaptations~\cite{yin2025causvid,huang2025selfforcing,zhu2026causal} have taken a promising step by converting bidirectional DiTs into causal, chunk-wise generators via asymmetric distillation~\cite{yin2025causvid} and self-forcing rollouts~\cite{huang2025selfforcing}, enabling real-time streaming at $480$P on a single GPU. However, scaling these methods to high resolutions (\eg, 1K or 2K) for practical deployment remains an open problem.

Since directly generating high-resolution video is prohibitively expensive, the cascaded paradigm~\cite{gao2025seedance,zhang2025flashvideo,team2025longcat}---first generating low-resolution video to capture semantics and motion, then upscaling  via super-resolution (SR) to supplement high-frequency detail---has emerged as a practical solution. However, existing cascaded approaches suffer from several fundamental limitations. Some streaming diffusion-based SR methods~\cite{zhuang2025flashvsr,shiu2025streamdiffvsr} achieve high efficiency but operate in pixel space, introducing additional encode--decode overhead~\cite{zhang2025waver} in the cascaded pipeline and requiring fundamental architectural modifications that forfeit the generative capability of the pre-trained T2V model, making training difficult. Other works attempt to address this through latent-space upsampling followed by a cascaded SR model~\cite{fsvideo2026,zhang2025flashvideo}. However, their upsampling strategies either adopt naive interpolation~\cite{zhang2025flashvideo,sii2026davinci} or rely on large-scale upsampler models~\cite{wu2025hunyuanvideo15,hacohen2026ltx2}. Both neglect the sensitivity of latent video representations to spatiotemporal consistency and cannot perform causal streaming extrapolation. Consequently, the subsequent SR stage must inject heavy noise to mitigate the frequency aliasing and spatiotemporal incoherence introduced by upsampling~\cite{fsvideo2026}. This heavy noise coverage over low-resolution information makes subsequent SR training---and its acceleration via distillation---considerably more difficult. Furthermore, existing high-resolution methods suffer from quadratic attention complexity and their SR components are not designed for one-step inference, making cascaded end-to-end optimization infeasible and leading to train-test inconsistency that compounds quality degradation.

\begin{figure}[t]
  \centering
  \includegraphics[width=.99\textwidth]{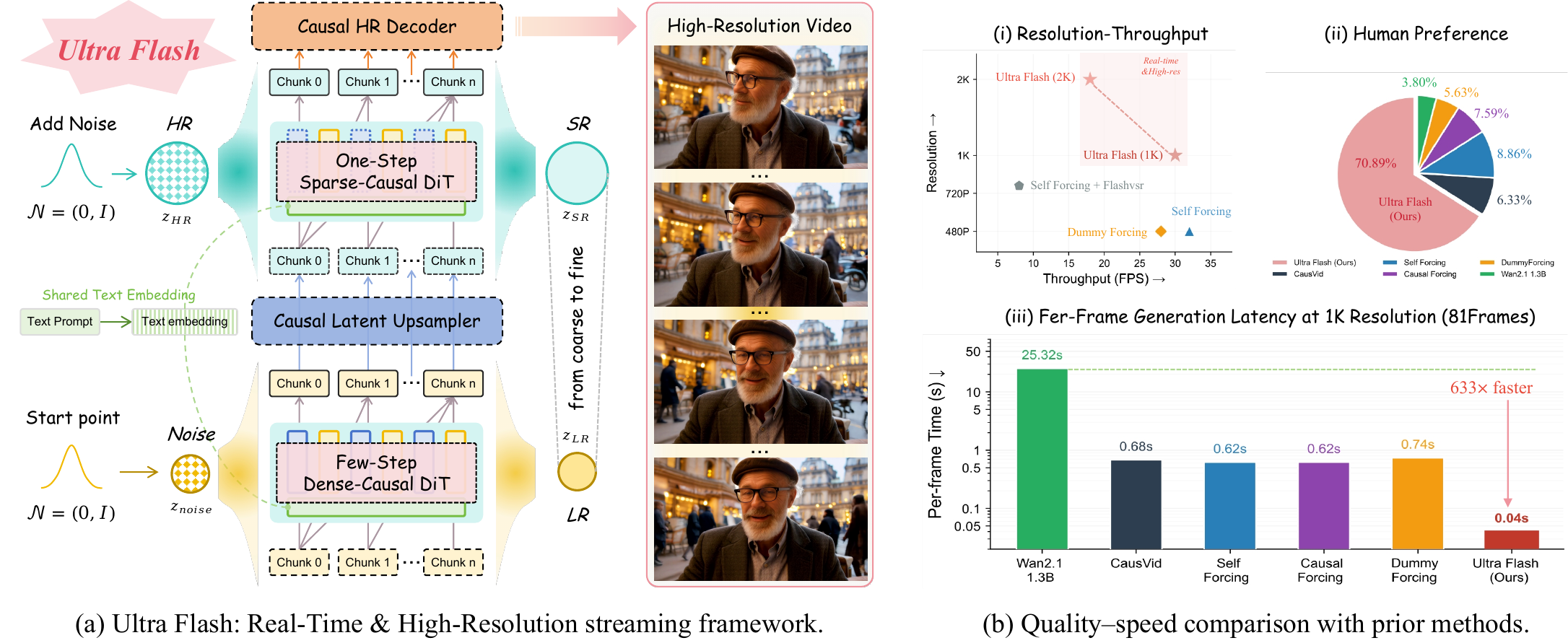}
  \caption{(a) \textbf{\method{}} framework. (b) Quality–speed comparison with prior methods. \method{} scales to 1K and 2K resolution while achieving better quality and real-time throughput
  }
  \label{fig:intro}
\end{figure}

To bridge this gap, we present \method{} (Fig.~\ref{fig:intro}), a cascaded streaming framework that scales real-time autoregressive video generation to high resolutions. \method{} achieves ${\sim}30$ FPS at 1K resolution and ${\sim}18$ FPS at 2K resolution on a single GPU through three key contributions:

\textbf{Efficient architecture-preserving T2V-to-TV2V SR training paradigm.} We propose an efficient training paradigm that converts any pre-trained T2V model into a TV2V multimodal generative SR model without architectural modification~\cite{zhuang2025flashvsr,shiu2025streamdiffvsr}, preserving the original generative capability. We further design an AIGC-oriented data degradation pipeline tailored to the characteristics of AI-generated video, effectively retaining model priors and enabling enhanced high-resolution detail when cascaded after mainstream low-resolution generative models.

\textbf{Ultralight streaming latent upsampler with high-resolution decoder.} 
  We design a causal memory network that upsamples low-resolution latents to high resolution directly in latent space with temporal coherence. Unlike pixel-space VSR methods~\cite{he2024venhancer,zhuang2025flashvsr} that introduce substantial overhead or latent cascaded approaches~\cite{zhang2025flashvideo,wu2025hunyuanvideo15} that rely on naive interpolation and restoration requiring heavy noise to mitigate aliasing, our spatiotemporally coherent upsampler adds $<$5\% pipeline cost while eliminating the need for heavy noise injection, substantially reducing SR training and distillation difficulty. Paired with a ultralight high-resolution decoder, Ultra Flash enables efficient latent spatial scaling and precise high-resolution decoding, laying the foundation for high-resolution streaming generation.

\textbf{Cascaded high-resolution streaming generation optimization.} Building on the above models, we devise a comprehensive optimization scheme to enable real-time high-resolution streaming. First, we perform \emph{hybrid-reward-enhanced sparse causalization and single-step distillation}: dynamic block-sparse causal attention~\cite{blocksparse2025} replaces dense attention for streaming-compatible inference, while distribution matching distillation~\cite{liu2025decoupledDMD} compresses multi-step denoising to a single step, with perceptual and aesthetic reward signals~\cite{xu2024refl,ke2021musiq,zhang2026omninft} to directly optimize for visual quality. Then, we introduce \emph{cascaded streaming self-forcing preference optimization with dynamic cache management}: the low-resolution generator and the high-resolution SR model are jointly rolled out in a cascaded streaming fashion, where a preference optimization objective explicitly trains on self-generated context to close the train-test gap, while a dynamic cache management mechanism can further enhance the generation efficiency. Together, these designs jointly enhance overall temporal coherence, improve visual quality, and realize real-time high-resolution streaming video generation.

\section{Method}
\label{sec:method}

\begin{figure}[!t]
  \centering
  \includegraphics[width=.99\textwidth]{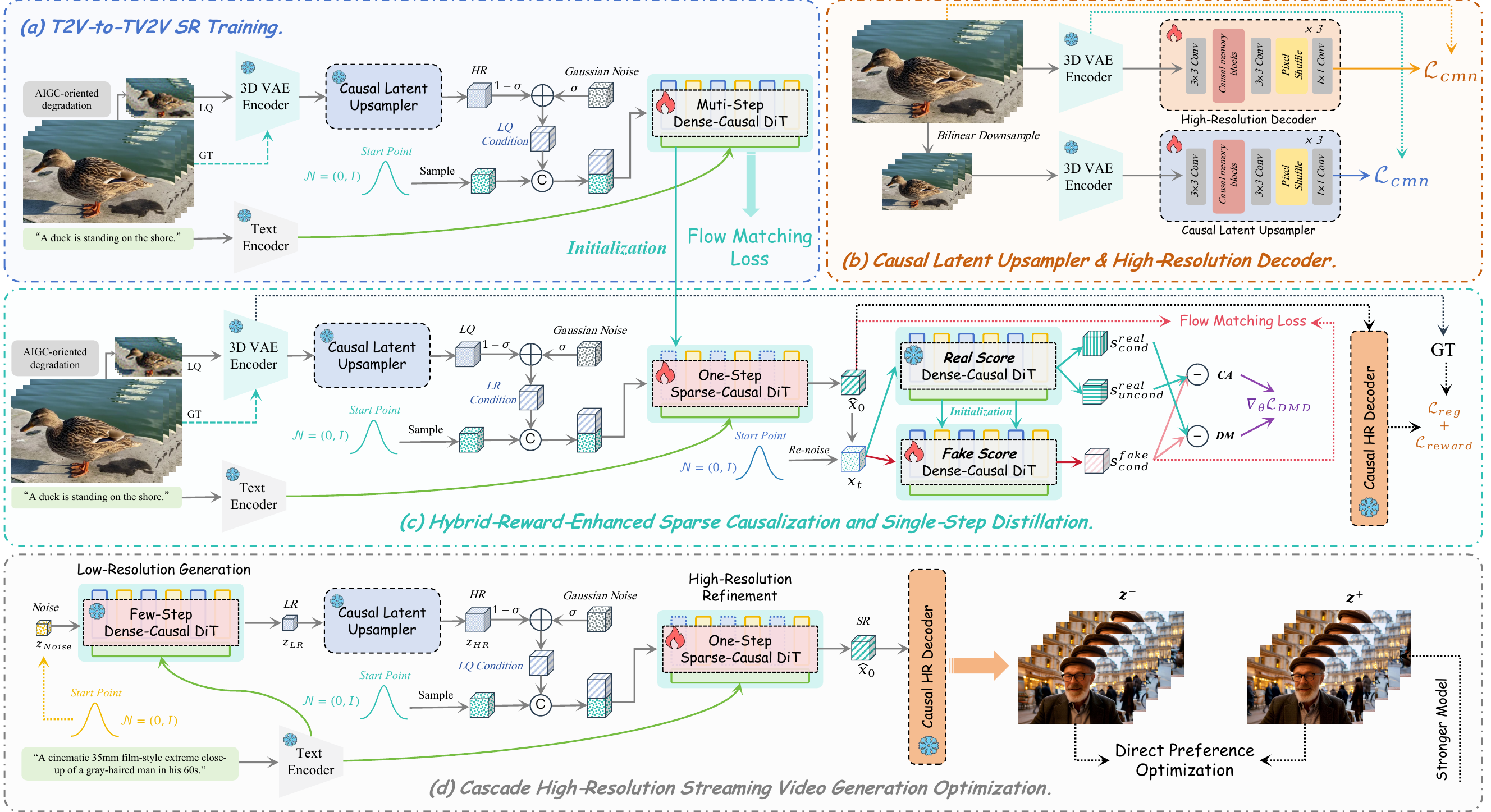}
  \caption{\textit{Detailed components and training of our Ultra Flash framework. (Zoom in for details.)} }
  \label{fig:pipeline}
\end{figure}

Given any low-resolution autoregressive streaming generator (\eg, Self Forcing~\cite{huang2025selfforcing}), \method{} cascades three components to scale real-time video generation to high resolutions (Fig.~\ref{fig:pipeline}): (1)~an \emph{architecture-preserving T2V-to-TV2V SR training paradigm} with an AIGC-oriented degradation pipeline that converts the pre-trained T2V model into a generative super-resolution model without architectural modification~(\S\ref{sec:sr_training}); (2)~a \emph{causal streaming latent upsampler} paired with a high-resolution decoder that lifts low-resolution latents to high resolution with spatiotemporal coherence~(\S\ref{sec:upsampler}); and (3)~a \emph{cascaded high-resolution streaming optimization} scheme comprising sparse causalization, single-step distillation, self-forcing preference optimization, and dynamic cache management~(\S\ref{sec:streaming_opt}).

\subsection{Architecture-Preserving T2V-to-TV2V SR Training Paradigm}
\label{sec:sr_training}

Existing pixel-space SR methods~\cite{zhuang2025flashvsr,shiu2025streamdiffvsr} require fundamental architectural modifications (\eg, LQ projection layers, modified attention patterns) that forfeit the generative capability of the pre-trained T2V model. We instead propose a paradigm that repurposes any T2V model as a TV2V generative SR model \emph{without architectural change}, preserving the full generative prior.

\noindent\textbf{Conditioning Mechanism.}
The upsampled low-resolution latent $\mathbf{z}^{\text{HR}}$ from the streaming latent upsampler is concatenated with the noise latent $\boldsymbol{\epsilon} \in \mathbb{R}^{t \times 2h \times 2w \times c}$ along the channel dimension, yielding a $2c$-channel input. The DiT's input projection is extended from $c$ to $2c$ channels, with the new weights initialized to zero so that training begins from the original T2V checkpoint. This zero-initialization preserves the model's generative capability at the start of training, and the model gradually learns to leverage the LR condition as training progresses.
To further enhance robustness and preserve generative capacity, we apply two conditioning augmentation strategies during training:
\emph{\textbf{(i)~Condition noise injection}}: Gaussian noise at a random level $\sigma_{\text{cond}} \in [\sigma_{\min}, \sigma_{\max}]$ is added to the LR condition latent before concatenation, preventing the model from overly relying on the condition and encouraging it to leverage its learned generative priors to complement missing detail.
\emph{\textbf{(ii)~Condition dropout}}: with probability $p_{\text{drop}}$, the LR condition is entirely zeroed out, forcing the model to perform pure T2V generation without any visual condition. This ensures the SR model retains strong unconditional generative capability, which both improves classifier-free guidance effectiveness and prevents mode collapse onto the LR input.

\noindent\textbf{AIGC-Oriented Data Degradation Pipeline.}
Standard degradation models designed for natural video~\cite{he2024venhancer} (\eg, Real-ESRGAN~\cite{wang2021realesrgan}) are poorly suited for AI-generated content, which exhibits characteristic artifacts distinct from natural camera noise---temporal flickering, unnatural motion jitter, rolling-shutter wobble, and diffusion-specific color shifts. We design a hybrid two-stage degradation pipeline that combines AIGC-specific temporal degradation with classical spatial degradation:

\emph{\textbf{Stage 1---AIGC synthetic degradation.}} A dedicated module applies five temporally coherent operations whose parameters evolve smoothly over time via low-frequency sinusoidal trajectories (avoiding inter-frame flicker):
\emph{(a)~Temporal morphing}: adjacent frames are alpha-blended with a time-varying mixing coefficient $\alpha_t \in [0.2, 0.9]$, simulating the exposure fusion artifacts common in diffusion-generated video.
\emph{(b)~Stochastic frame dropping}: frames are randomly dropped (probability $p_{\text{drop}}$ with a maximum consecutive-drop constraint) and reconstructed via linear interpolation, emulating temporal jitter from autoregressive generation.
\emph{(c)~Directional motion blur}: per-frame line kernels with temporally smooth angle $\theta_t$ and length $l_t$ produce spatially varying motion blur, mimicking the anisotropic blur patterns unique to diffusion denoising.
\emph{(d)~ROI-constrained grid warping}: a low-frequency displacement field is generated and masked by a temporally drifting soft-ellipse ROI, producing localized geometric distortion that resembles rolling-shutter wobble in AI-generated videos.
\emph{(e)~Video codec compression}: H.264 encoding at randomized CRF levels introduces block artifacts and quantization noise characteristic of compressed AIGC outputs.

\emph{\textbf{Stage 2---Cascaded spatial degradation.}} Following the AIGC stage, a Real-ESRGAN-style~\cite{wang2021realesrgan} two-pass spatial degradation is applied: each pass consists of USM sharpening, Gaussian blur (kernel size $\in [15, 37]$, $\sigma \in [0.2, 3.0]$), random rescaling (factor $\in [0.15, 1.5]$), additive noise, and JPEG compression ($q \in [70, 95]$). The two stages are cascaded to produce diverse, realistic degradation. Finally, a $2{\times}$--$4{\times}$ bicubic downsampling followed by upsampling back to the original resolution simulates the spatial resolution gap. A stochastic mixing strategy (CutMix between the AIGC-degraded and spatially-degraded branches) further enriches training diversity.

This hybrid pipeline generates realistic LR--HR training pairs that faithfully simulate the degradation characteristics of AI-generated video, enabling the SR model to effectively restore AIGC-specific artifacts while retaining the base model's generative priors.

\noindent\textbf{SR Training Objective.}
The SR model is trained with the standard flow matching objective~\cite{liu2022flow}. Given the clean HR latent $\mathbf{z}_0$ and sampled noise $\boldsymbol{\epsilon}$, we construct the noisy latent $\mathbf{z}_t = (1-\sigma_t)\mathbf{z}_0 + \sigma_t \boldsymbol{\epsilon}$ at timestep $t$, where $\sigma_t$ is drawn from a log-normal distribution with a shifted schedule (flow shift $s$). The model $f_\theta$ predicts the velocity field, trained via:
\begin{equation}
  \mathcal{L}_{\text{FM}} = \mathbb{E}_{t, \boldsymbol{\epsilon}} \left\| f_\theta(\mathbf{z}_t,\, t,\, \mathbf{c}_{\text{text}},\, \mathbf{z}^{\text{HR}}) - (\boldsymbol{\epsilon} - \mathbf{z}_0) \right\|_2^2,
  \label{eq:reg}
\end{equation}
where the conditioning $\mathbf{z}^{\text{HR}}$ (upsampled LR latent) is concatenated along the channel dimension and the text prompt $\mathbf{c}_{\text{text}}$ provides semantic guidance. Combined with the condition noise injection and dropout described above, this training scheme enables classifier-free guidance (CFG) at inference, where the model can be steered between conditional SR and unconditional generation. At inference, multi-step ODE integration with CFG yields high-quality HR outputs from the trained flow. This multi-step SR model serves as the teacher for the subsequent single-step distillation stage (\S\ref{sec:streaming_opt}).

\subsection{Causal Streaming Latent Upsampler with High-Resolution Decoder}
\label{sec:upsampler}

In cascaded high-resolution (HR) generation, the upsampling stage bridges low-resolution (LR) latents and the subsequent SR model. As discussed in \S\ref{sec:intro}, existing approaches either use naive interpolation---introducing frequency aliasing that forces the SR stage to inject heavy noise---or employ heavyweight upsampler models that preclude streaming. We propose a unified causal memory network architecture that serves as both the streaming latent upsampler and the high-resolution decoder, achieving configurable spatiotemporal upsampling and decoding with minimal overhead.

\noindent\textbf{Unified Causal Memory Network Architecture.}
Both the latent upsampler and the HR decoder share the same multi-stage architecture, differing only in their configured spatial/temporal scale factors and channel dimensions. An input $3{\times}3$ convolution followed by three cascaded stages, each consisting of $N_b$ causal memory blocks, a spatial upsampling layer, a temporal expansion layer, and a channel transition convolution. The core building block is the \emph{CausalMemBlock}, which fuses the current frame's feature with the memory from the previous frame:
\begin{equation}
  \mathbf{h}_t^{(\ell)} = \sigma\!\left( \text{Conv}_{3{\times}3}^{(3)}\!\left(\text{Conv}_{3{\times}3}^{(2)}\!\left(\sigma\!\left(\text{Conv}_{3{\times}3}^{(1)}\!\left([\mathbf{h}_t^{(\ell-1)};\; \mathbf{m}_{t-1}^{(\ell)}]\right)\right)\right)\right) + W_{\text{skip}} \mathbf{h}_t^{(\ell-1)} \right),
  \label{eq:memblock}
\end{equation}
where $[\cdot;\cdot]$ denotes channel concatenation, $\mathbf{m}_{t-1}^{(\ell)}$ is the previous frame's feature serving as temporal memory, $\sigma$ is ReLU activation, and $W_{\text{skip}}$ is a $1{\times}1$ projection for residual connection. The causal structure ensures each frame depends only on past context, enabling frame-wise streaming inference.

Spatial upsampling within each stage is performed via PixelShuffle~\cite{shi2016pixelshuffle}---a $3{\times}3$ convolution expands the channel dimension by $r^2$ followed by a sub-pixel rearrangement that converts channels into spatial resolution (per-stage factor of $1{\times}$ or $2{\times}$ or $3{\times}$). Temporal upsampling is realized via a \emph{temporal expansion} operator---a $1{\times}1$ convolution that lifts the channel dimension by factor $s$, followed by a channel-to-time reshape that unfolds $s$ new frames from each input frame. By configuring stage factors independently, the same architecture supports $n{\times}$ spatial upsampling for the latent upsampler ($n=a\times b \times c$, spatial\_factors$=[a,b,c]$, temporal\_factors$=[1,1,1]$) and $8{\times}8{\times}4$ spatiotemporal decoding for the HR decoder (spatial\_factors$=[2,2,2]$, temporal\_factors$=[1,2,2]$).

\noindent\textbf{Optimization.}
Both streaming latent upsampler and HR decoder are trained with an MSE reconstruction loss combined with an optical-flow-warped temporal consistency (eWarp) loss:
\begin{equation}
  \mathcal{L}_{\text{cmn}} = \left\| \hat{\mathbf{x}} - \mathbf{x}_{\text{gt}} \right\|_2^2 + \lambda_{\text{warp}} \cdot \mathcal{L}_{\text{eWarp}},
  \label{eq:upsampler_loss}
\end{equation}
where $\hat{\mathbf{x}}$ denotes the network output---latent features $\hat{\mathbf{z}}$ for the upsampler or decoded pixels $\hat{\mathbf{I}}$ for the HR decoder---and $\mathcal{L}_{\text{eWarp}} = \sum_{t} \| \hat{\mathbf{x}}_t - \text{Warp}(\hat{\mathbf{x}}_{t-1}, \mathbf{F}_{t \to t-1}) \|$ penalizes temporal inconsistency by warping adjacent frames using the estimated optical flow $\mathbf{F}$. This encourages spatiotemporally smooth outputs, directly reducing the noise level required by the downstream SR model.

\subsection{Cascaded High-Resolution Streaming Generation Optimization.}
\label{sec:streaming_opt}

The multi-step SR model from \S\ref{sec:sr_training} produces high-quality results but relies on dense bidirectional attention with iterative denoising, far from real-time. We devise a two-phase optimization scheme that progressively transforms it into a real-time streaming model: \textbf{\emph{Phase~I}} converts the SR model into a single-step causal generator via hybrid-reward-enhanced sparse causalization and distillation; \textbf{\emph{Phase~II}} closes the train-test gap of the full cascaded pipeline via self-forcing preference optimization, coupled with dynamic cache management that further improves inference efficiency.

\subsubsection{Hybrid-Reward-Enhanced Sparse Causalization and Single-Step Distillation}
\label{sec:sparse_distill}

This phase addresses two orthogonal bottlenecks simultaneously---the dense bidirectional attention precludes streaming, and multi-step denoising dominates latency---while injecting perceptual reward signals to compensate for the quality degradation typically incurred by acceleration.

\noindent\textbf{Dynamic Block-Sparse Causal Attention.}
To enable streaming-compatible inference, we replace the dense bidirectional attention with \emph{dynamic block-sparse causal attention}. The 3D token grid $(t, h, w)$ is divided into non-overlapping blocks of $(b_t, b_h, b_w)$, yielding $N_b$ blocks. For each attention layer, a block-level mask is computed in two stages:
\emph{\textbf{(a)~Structural masks:}} a spatial locality mask $\mathbf{M}_{\text{local}} \in \{0,1\}^{N_b \times N_b}$ confines each block's receptive field to a sliding window of size $r {\times} r$, and a temporal causal mask $\mathbf{M}_{\text{causal}}$ ensures each temporal chunk attends only to current and preceding chunks.
\emph{\textbf{(b)~Content-adaptive top-$k$ selection:}} we pool Q and K within each block, compute block-level attention scores, and retain the top-$k$ most relevant blocks per head:
\begin{equation}
  s_{ij}^h = \frac{\bar{\mathbf{q}}_i^{h\top} \bar{\mathbf{k}}_j^h}{\sqrt{d_h}}, \quad
  \mathbf{M}_{\text{sparse}}^h[i, j] = \mathbb{1}\!\left[ \text{softmax}(s_{i,:}^h)_j \geq \tau_k \right] \cap \mathbf{M}_{\text{local}} \cap \mathbf{M}_{\text{causal}},
  \label{eq:sparse_mask}
\end{equation}
where $h$ indexes the attention head, $\bar{\mathbf{q}}_i, \bar{\mathbf{k}}_j$ are block-mean pooled queries and keys, $d_h$ is the per-head dimension, $\tau_k$ is the adaptive threshold corresponding to the top-$k$ budget, and $\mathbb{1}[\cdot]$ is the indicator function. The per-head mask is passed to a block-sparse attention kernel~\cite{blocksparse2025} for hardware-efficient execution, with the top-$k$ budget scaling adaptively with resolution to maintain consistent sparsity.

\noindent\textbf{Single-Step Distillation via Decoupled DMD.}
We adopt Decoupled DMD~\cite{liu2025decoupledDMD} to compress multi-step denoising into a single forward pass. The distillation involves three models: \emph{(i)}~the \emph{real score model} (teacher)---the multi-step bidirectional SR model with CFG trained in \S\ref{sec:sr_training}, kept frozen; \emph{(ii)}~the \emph{fake score model}---initialized from the teacher weights and trained on student-generated samples via a flow matching objective to track the evolving generator distribution, updated at 5$\times$ the generator's frequency; and \emph{(iii)}~the \emph{generator} (student)---converted to causal sparse attention and being the primary trainable component. The decoupled DMD gradient decomposes into a CFG Augmentation (CA) term that drives the multi-step-to-single-step conversion, and a Distribution Matching (DM) regularizer that stabilizes generation quality:
\begin{equation}
  \nabla_{\theta} \mathcal{L}_{\text{d-DMD}} = \mathbb{E} \left[- \left( \underbrace{s^{\text{real}}_{\text{cond}}(\mathbf{x}_{\tau_{\text{DM}}}) - s^{\text{fake}}_{\text{cond}}(\mathbf{x}_{\tau_{\text{DM}}})}_{\text{DM regularizer}} + (\alpha{-}1) \underbrace{\left(s^{\text{real}}_{\text{cond}}(\mathbf{x}_{\tau_{\text{CA}}}) - s^{\text{real}}_{\text{uncond}}(\mathbf{x}_{\tau_{\text{CA}}})\right)}_{\text{CA engine}}  \right) \frac{\partial G_{\theta}}{\partial\theta} \right],
  \label{eq:dmd}
\end{equation}
where $s^{\text{real}}_{\text{cond/uncond}}$ and $s^{\text{fake}}_{\text{cond}}$ denote the conditional/unconditional score predictions from the real and fake models respectively, $\alpha$ is the CFG scale, $G_\theta$ is the student generator, and $\tau_{\text{CA}} > t$, $\tau_{\text{DM}} \in [0,1]$ are decoupled re-noising schedules. Since the SR task has access to ground-truth HR targets, we further introduce a wavelet L1 loss (omitting the LL sub-band to emphasize high-frequency detail) and an LPIPS perceptual loss to constrain pixel-level reconstruction via the HR decoder:
\begin{equation}
  \mathcal{L}_{\text{reg}} = \lambda_{\text{wav}} \left\| \mathcal{W}_{\text{HF}}(\mathcal{D}_{\text{HR}}(\hat{\mathbf{z}}_0)) - \mathcal{W}_{\text{HF}}(\mathbf{I}_{\text{gt}}) \right\|_1 + \lambda_{\text{lpips}} \cdot \text{LPIPS}\!\left(\mathcal{D}_{\text{HR}}(\hat{\mathbf{z}}_0),\, \mathbf{I}_{\text{gt}}\right),
  \label{eq:pixel_reg}
\end{equation}
where $\mathcal{W}_{\text{HF}}$ denotes the high-frequency wavelet sub-bands (LH, HL, HH) and $\mathcal{D}_{\text{HR}}$ is the differentiable HR decoder enabling gradient back-propagation to the student.

\noindent\textbf{Hybrid Reward Integration.}
Beyond reconstruction losses, we integrate perceptual and aesthetic reward signals from frozen quality predictors to directly optimize the student's visual quality:
\begin{equation}
  \mathcal{L}_{\text{reward}} = -\lambda_{\text{clip}} \cdot \text{CLIP-IQA}^+\!\left(\mathcal{D}_{\text{HR}}(\hat{\mathbf{z}}_0)\right) - \lambda_{\text{musiq}} \cdot \text{MUSIQ}\!\left(\mathcal{D}_{\text{HR}}(\hat{\mathbf{z}}_0)\right) - \lambda_{\text{aes}} \cdot \text{LAION-Aes}\!\left(\mathcal{D}_{\text{HR}}(\hat{\mathbf{z}}_0)\right),
  \label{eq:iqa}
\end{equation}
where CLIP-IQA$^+$~\cite{wang2023clipiqa} captures perceptual quality, MUSIQ~\cite{ke2021musiq} evaluates multi-scale image quality, and the LAION-Aesthetic predictor~\cite{schuhmann2022laion} assesses aesthetic appeal. Gradients flow through $\mathcal{D}_{\text{HR}}$ back to the student, providing complementary signals that directly enhance sharpness, color fidelity, and visual aesthetics beyond what distribution matching alone achieves. The SR model of student's total training objective in Phase~I is:
\begin{equation}
  \mathcal{L}_{\text{Phase\,I}} = \mathcal{L}_{\text{d-DMD}} + \mathcal{L}_{\text{reg}} + \mathcal{L}_{\text{reward}}.
  \label{eq:phase1_loss}
\end{equation}

\subsubsection{Cascaded Streaming Self-Forcing Preference Optimization and Cache Management}
\label{sec:self_forcing}

Phase~I produces a single-step causal SR model, but it is still trained on ground-truth low-resolution context. At inference, the SR model must instead operate on imperfect outputs from the upstream streaming generator and its own prior predictions---a compounded exposure bias unique to the cascaded setting. Phase~II addresses this via joint cascaded rollout with preference optimization.

\noindent\textbf{High-Resolution Self-Forcing Rollout.}
During training, we simulate the actual inference distribution by performing high-resolution rollout of the entire cascaded pipeline: the low-resolution streaming generator produces context chunks autoregressively, which are upsampled by the latent upsampler and fed to the single-step SR model. The SR output in turn serves as context for subsequent chunks, exposing the model to its own imperfections and upstream errors simultaneously. By adjusting the spatial upsampling factor of the latent upsampler, we flexibly support both 1K and 2K streaming generation.

\noindent\textbf{Preference Optimization.}
We apply Direct Preference Optimization (DPO)~\cite{rafailov2023dpo} to the entire cascaded streaming pipeline, updating only the SR model's parameters. Preference pairs are constructed as follows: \emph{negative samples} $\mathbf{z}^{-}$ are generated by the current cascaded pipeline in streaming mode (single-step causal SR), while \emph{positive samples} $\mathbf{z}^{+}$ are produced by a stronger Wan2.2-5B SR model performing multi-step pixel-space super-resolution, serving as an oracle reference. The DPO loss directly optimizes the SR model to shift its output distribution toward the higher-quality reference:
\begin{equation}
  \mathcal{L}_{\text{Phase\,II}} = \mathcal{L}_{\text{pref}} = -\log \sigma\!\left( \beta \left( \log \frac{\pi_\theta(\mathbf{z}^{+} | \mathbf{c})}{\pi_{\text{ref}}(\mathbf{z}^{+} | \mathbf{c})} - \log \frac{\pi_\theta(\mathbf{z}^{-} | \mathbf{c})}{\pi_{\text{ref}}(\mathbf{z}^{-} | \mathbf{c})} \right) \right),
  \label{eq:pref}
\end{equation}
where $\pi_\theta$ is the SR model being optimized, $\pi_{\text{ref}}$ is the frozen reference policy (the Phase~I checkpoint), $\mathbf{c}$ denotes the conditioning context from the upstream cascaded pipeline, and $\beta$ is the temperature.

\noindent\textbf{Dynamic Cache Management.}
At inference, we apply three complementary strategies to reduce per-chunk computation across the cascaded pipeline.
\emph{\textbf{(i)~LR denoising step reduction.}} The upstream LR streaming generator (\eg, Self Forcing~\cite{huang2025selfforcing}) nominally uses 4 denoising steps per chunk. Since the downstream SR model compensates for fine detail, the LR output does not require full fidelity. We therefore run the complete 4-step schedule only for the first chunk and reduce all subsequent chunks to 3 steps, saving one forward pass per chunk without perceptible quality degradation.
\emph{\textbf{(ii)~Adaptive cache refresh.}} By default, after the denoising steps the generator performs an additional forward pass on the predicted $\mathbf{x}_0$ to compute fresh KV entries for the cache. We instead evaluate the previous chunk's decoded frames with a lightweight IQA metric: if the score exceeds a predefined threshold, we directly reuse the KV cache from the last denoising step $\mathbf{x}_t$, eliminating the extra forward pass and further reducing latency.
\emph{\textbf{(iii)~SR cache length adaptation.}} For the single-step SR model, the strong conditioning signal from the upsampled LR latent makes generation quality relatively insensitive to KV cache length. We exploit this by dynamically selecting a compact cache window, trading minimal quality for significant memory and compute savings during extended sequence generation.

\begin{table}[!t]
  \centering
  \caption{\textbf{VBench comparison.} SC: subject consistency; BC: background consistency; MS: motion smoothness; IQ: imaging quality; AQ: aesthetic quality. $\uparrow$: higher is better.}
  \label{tab:vbench}
  \small
  \begin{tabular}{lccccccc}
    \toprule
    Method & Steps & SC~$\uparrow$ & BC~$\uparrow$ & MS~$\uparrow$ & IQ~$\uparrow$ & AQ~$\uparrow$ & Total~$\uparrow$ \\
    \midrule
    Wan2.1 & 50 & 96.87 & 97.35 & 98.31 & 65.02 & 62.79 & 83.37 \\
    CausVid~\cite{yin2025causvid} & 4 & 96.58 & 96.74 & 97.82 & 67.45 & 63.87 & 81.33 \\
    Self Forcing~\cite{huang2025selfforcing} & 4 & 97.12 & 97.08 & 97.63 & 67.92 & 63.50 & 83.74 \\
    Causal Forcing~\cite{zhu2026causal} & 4 & 97.53 & 97.19 & 98.05 & 68.88 & 64.15 & 84.04 \\
    DummyForcing~\cite{guo2026dummy} & 4 & 96.45 & 96.81 & 98.14 & 67.28 & 63.62 & 83.48 \\
    \midrule
    {\method{} (Ours)} & 4 & {97.68} & {97.42} & {98.13} & {68.90} & {64.21} & {84.17} \\
    \bottomrule
  \end{tabular}
\end{table}
 
\begin{table}[!t]
  \centering
  \caption{\textbf{Efficiency comparison.} Throughput (FPS), per-frame latency, and peak GPU memory for streaming generation. All measurements on a single GPU.}
  \label{tab:efficiency}
  \resizebox{\textwidth}{!}{
  \begin{tabular}{lcccccc}
    \toprule
    Method & Resolution & Steps & FPS~$\uparrow$ & Latency (ms)~$\downarrow$ & Streaming \\
    \midrule
    Wan2.1~\cite{wan2025wan} & $480{\times}832$ & 50 & 0.78 & 103,000 & $\times$ \\
    CausVid~\cite{yin2025causvid} & $480{\times}832$ & 4 & 29.4 & 34 & \checkmark \\
    Self Forcing~\cite{huang2025selfforcing} & $480{\times}832$ & 4 & 32.0 & 31 & \checkmark \\
    Causal Forcing~\cite{zhu2026causal} & $480{\times}832$ & 4 & 31.2 & 32 & \checkmark \\
    DummyForcing~\cite{guo2026dummy} & $480{\times}832$ & 4 & 28.0 & 36 & \checkmark \\
    Self Forcing + FlashVSR~\cite{zhuang2025flashvsr} & $768{\times}1408$ & 5 & 15.0 & 67 & \checkmark \\
    \midrule
    {\method{} (Ours)} & $960{\times}1664$ (1K) & 4 & {30.0} & {40} & \checkmark \\
    {\method{} (Ours)} & $1440{\times}2496$ (2K) & 4 & 18.0 & 56 & \checkmark \\
    \bottomrule
  \end{tabular}
  }
\end{table}

\begin{table}[!t]
  \centering
  \caption{\textbf{SR quality comparison.} We compare upsampling strategies (interpolation \textit{vs.}\ streaming latent upsampler) and optimization stages (multi-step \textit{vs.}\ distilled \textit{vs.}\ full pipeline).}
  \label{tab:sr}
  \small
  \begin{tabular}{lcccc}
    \toprule
    Configuration & CLIP-IQA+~$\uparrow$ & MUSIQ~$\uparrow$ & NIQE~$\downarrow$ & FPS~$\uparrow$ \\
    \midrule
    Bilinear + Multi-step SR & 0.612 & 68.3 & 4.82 & 1.8 \\
    Bicubic + Multi-step SR & 0.625 & 69.1 & 4.65 & 1.8 \\
    Upsampler + Multi-step SR & 0.671 & 72.4 & 4.18 & 1.6 \\
    Upsampler + Sparse single-step & 0.658 & 71.2 & 4.31 & 30.0 \\
    \midrule
    {Full pipeline (+ pref.\ opt.)} & {0.692} & {73.8} & {3.95} & {30.0} \\
    \bottomrule
  \end{tabular}
\end{table}

\begin{figure}[ht]
  \centering
  \includegraphics[width=\textwidth]{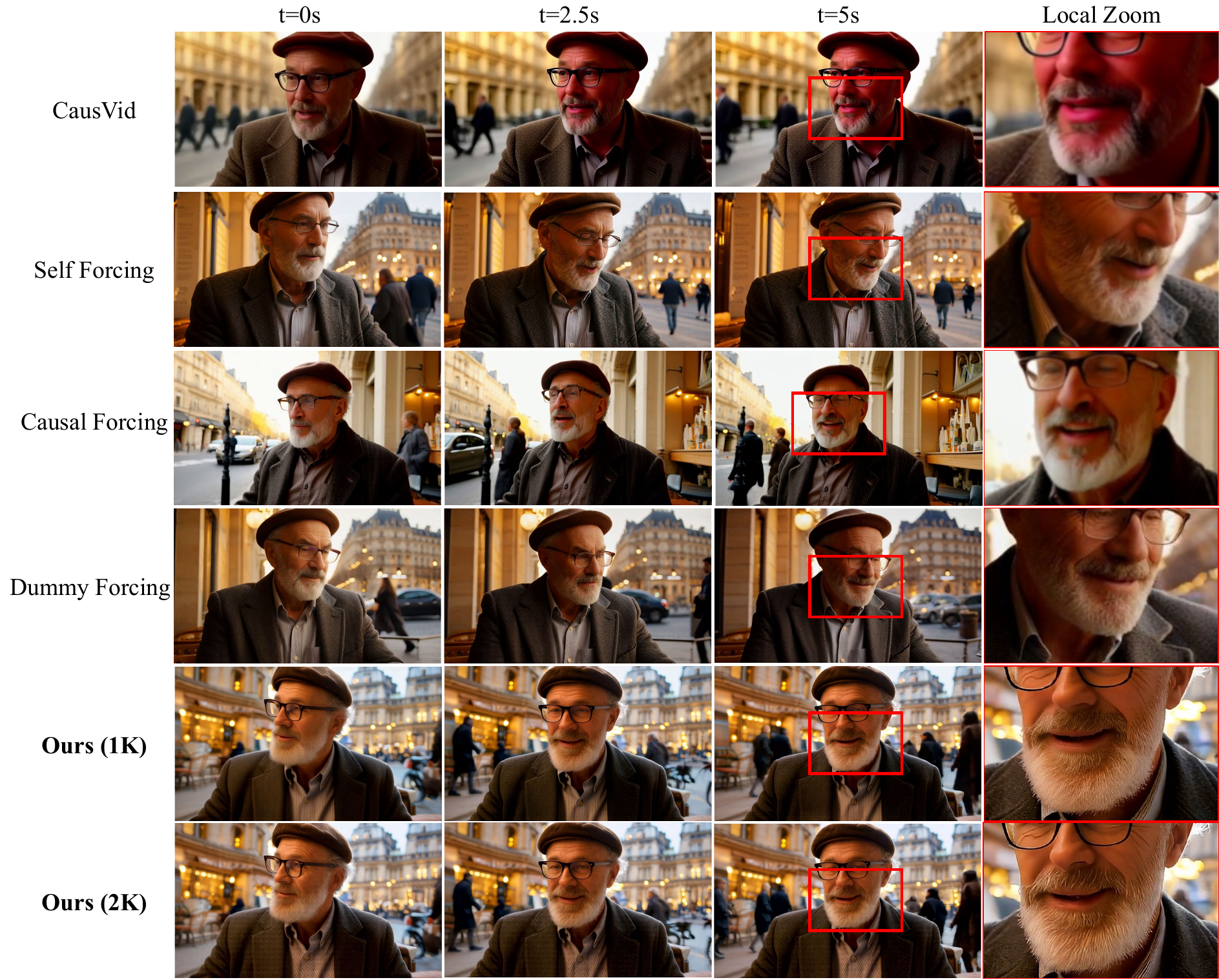}
  \caption{\textbf{Qualitative comparison.} \method{} real-time generates sharper, temporally coherent frames at 1K \& 2K while prior methods are limited to $480{\times}832$.  
  \textit{\textbf{Zoom in for the details.}}
  }
  \label{fig:qualitative2}
\end{figure}

\begin{figure}[ht]
  \centering
  \includegraphics[width=\textwidth]{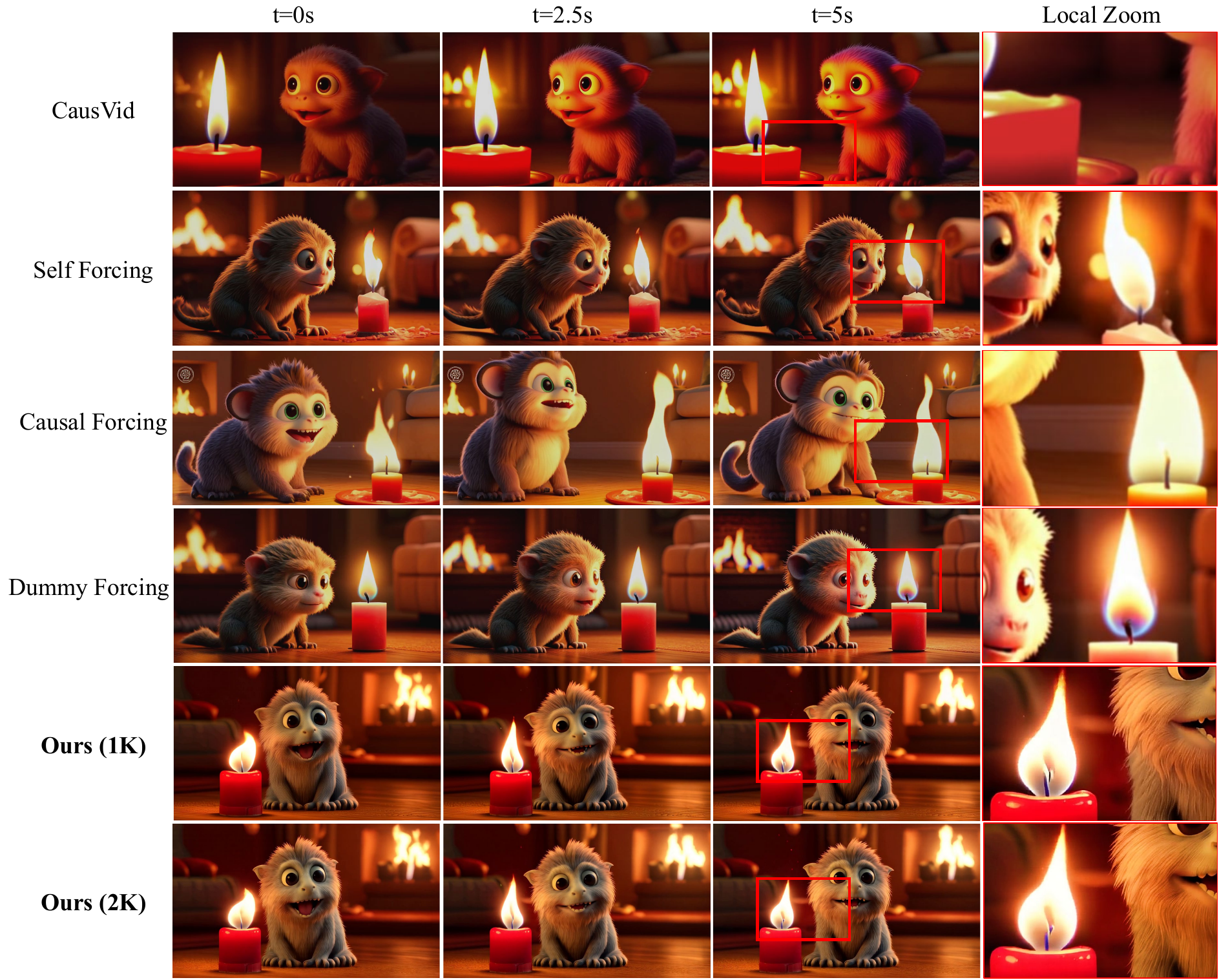 }
  \caption{\textbf{Qualitative comparison.} \method{} real-time generates sharper, temporally coherent frames at 1K \& 2K while prior methods are limited to $480{\times}832$.  
  \textit{\textbf{Zoom in for the details.}}
  }
  \label{fig:qualitative3}
\end{figure}

\begin{figure}[ht]
  \centering
  \includegraphics[width=\textwidth]{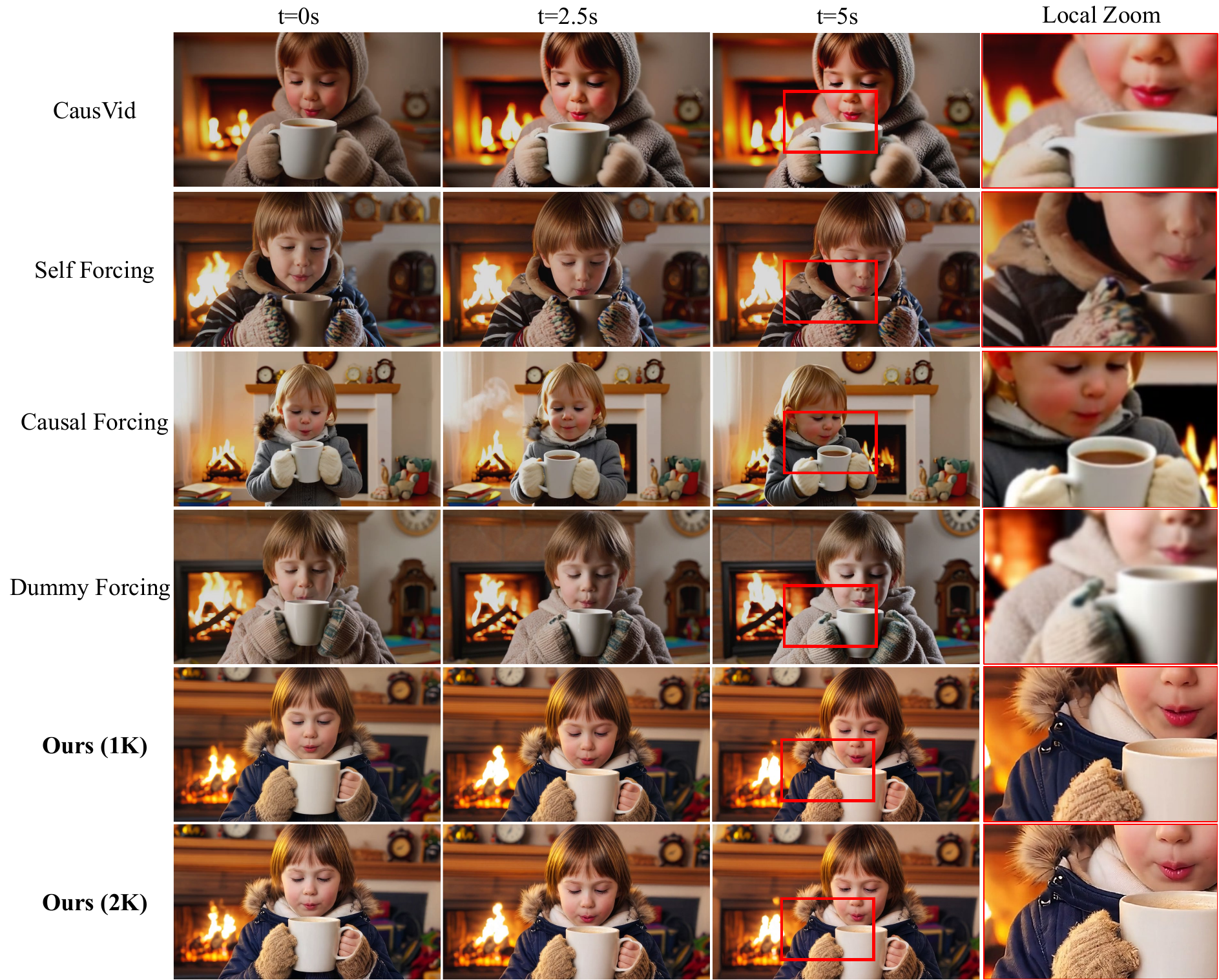 }
  \caption{\textbf{Qualitative comparison.} \method{} real-time generates sharper, temporally coherent frames at 1K \& 2K while prior methods are limited to $480{\times}832$.  
  }
  \vspace{-2mm}
  \label{fig:qualitative4}
\end{figure}

\section{Experiments}
\label{sec:experiments}

\subsection{Experimental Setup}
\label{sec:setup}

\noindent\textbf{Base Model and Pipeline.}
We build \method{} on top of Wan2.1-T2V-1.3B~\cite{wan2025wan}, a 1.3B-parameter video DiT with 30 transformer blocks, hidden dimension 1536, 12 attention heads, and a UMT5-XXL text encoder. A pre-trained low-resolution streaming generator produces 480P latents, which are fed to the streaming latent upsampler, the generative SR model, and the HR decoder in cascade.

\noindent\textbf{Evaluation.}
We evaluate on \textbf{VBench}~\cite{huang2024vbench}, perceptual quality metrics (CLIP-IQA+, MUSIQ, NIQE), and efficiency metrics (FPS, latency per chunk) on a single H200/B200 GPU.

\noindent\textbf{Baselines.}
We compare with: (1)~\textbf{CausVid}~\cite{yin2025causvid}: autoregressive DMD with dense attention; (2)~\textbf{Self Forcing}~\cite{huang2025selfforcing}: self-rollout training for AR video; (3)~\textbf{Causal Forcing}~\cite{zhu2026causal}: improved AR distillation; (4)~\textbf{DummyForcing}~\cite{guo2026dummy}: dummy-head acceleration; (5)~\textbf{FlashVSR}~\cite{zhuang2025flashvsr}: sparse-attention pixel-space streaming SR. All streaming baselines operate at 480P; FlashVSR operates at $768{\times}1408$ with an external LQ video input. \textit{\textbf{More details can be found in the appendix.}}

\subsection{Main Results}

\noindent\textbf{Quality Comparison.}
Table~\ref{tab:vbench} reports VBench scores. \method{} achieves competitive or superior quality to prior few-step methods on all dimensions while being the only method that additionally supports real-time high-resolution generation.

\noindent\textbf{Efficiency and Resolution Scaling.}
Table~\ref{tab:efficiency} compares efficiency across methods and resolutions. \method{} achieves ${\sim}$30 FPS at 1K ($960{\times}1664$) and ${\sim}$18 FPS at 2K ($1440{\times}2496$) on a single B200 GPU and ${\sim}$17 FPS \& ${\sim}$10 FPS on a single H200 GPU, significantly outperforming existing methods.

\noindent\textbf{High-Resolution SR Quality.}
Table~\ref{tab:sr} evaluates the generative SR quality at 1K resolution, comparing different upsampling strategies and optimization stages within the \method{} pipeline.

\begin{table}[!t]
  \centering
  \caption{\textbf{Ablation: SR training components.} Impact of AIGC degradation, zero-init conditioning, condition noise injection, and condition dropout.}
  \label{tab:ablation_sr_training}
  \small
  \begin{tabular}{lcccc}
    \toprule
    Configuration & CLIP-IQA+~$\uparrow$ & MUSIQ~$\uparrow$ & NIQE~$\downarrow$ & LPIPS~$\downarrow$ \\
    \midrule
    Full SR training & {0.671} & {72.4} & {4.18} & {0.182} \\
    $-$ AIGC degradation & 0.638 & 69.5 & 4.67 & 0.205 \\
    $-$ Zero-init conditioning & 0.621 & 68.1 & 4.83 & 0.218 \\
    $-$ Condition noise injection & 0.644 & 70.2 & 4.52 & 0.197 \\
    $-$ Condition dropout & 0.652 & 70.8 & 4.41 & 0.193 \\
    \bottomrule
  \end{tabular}
\end{table}

\begin{table}[!t]
  \centering
  \caption{\textbf{Ablation: upsampling strategy.} Impact on downstream SR quality and latency overhead.}
  \label{tab:ablation_upsample}
  \small
  \begin{tabular}{lcccc}
    \toprule
    Upsampler & CLIP-IQA+~$\uparrow$ & MUSIQ~$\uparrow$ & NIQE~$\downarrow$ & Overhead (ms) \\
    \midrule
    Bilinear interpolation & 0.612 & 68.3 & 4.82 & $<$1 \\
    Bicubic interpolation & 0.625 & 69.1 & 4.65 & $<$1 \\
    Causal memory (4 blocks) & 0.658 & 71.6 & 4.28 & 5 \\
    Causal memory (8 blocks) & {0.671} & {72.4} & {4.18} & 8 \\
    \bottomrule
  \end{tabular}
\end{table}

\begin{table}[!t]
  \centering
  \caption{\textbf{Ablation: cascaded streaming optimization components.} Each row removes one component from the full pipeline. TC: temporal consistency (VBench). Mem: peak GPU memory.}
  \label{tab:ablation_opt}
  \small
  \begin{tabular}{lccccc}
    \toprule
    Configuration & CLIP-IQA+~$\uparrow$ & MUSIQ~$\uparrow$ & TC~$\uparrow$ & FPS~$\uparrow$ & Mem (GB)~$\downarrow$ \\
    \midrule
    Full pipeline & {0.692} & {73.8} & {97.2} & {30.0} & {12.6} \\
    $-$ Adaptive top-$k$ (fixed window) & 0.674 & 72.1 & 96.8 & 28.5 & 12.6 \\
    $-$ Hybrid reward $\mathcal{L}_{\text{reward}}$ & 0.665 & 71.4 & 97.0 & 30.0 & 12.6 \\
    $-$ DPO pref.\ $\mathcal{L}_{\text{pref}}$ & 0.673 & 72.5 & 95.6 & 30.0 & 12.6 \\
    $-$ Dynamic cache management & 0.690 & 73.6 & 97.1 & 27.4 & 16.8 \\
    \bottomrule
  \end{tabular}
\end{table}

\subsection{Ablation Studies}
\label{sec:ablation}

We ablate the three core contributions of \method{} to validate each component. All ablations are conducted at $960{\times}1664$ unless stated otherwise.

\noindent\textbf{Architecture-Preserving SR Training.}
Table~\ref{tab:ablation_sr_training} validates the SR training paradigm. Removing the AIGC-oriented degradation pipeline (using standard Real-ESRGAN degradation only) reduces quality, confirming that tailored degradation effectively simulates AI-generated artifacts. Removing zero-initialized conditioning (replacing with random init) destabilizes early training. Disabling condition noise injection causes the model to overfit to the LR input and lose generative capability, while disabling condition dropout weakens CFG effectiveness.

\noindent\textbf{Causal Streaming Latent Upsampler.}
Table~\ref{tab:ablation_upsample} compares upsampling strategies. Naive interpolation introduces frequency aliasing that degrades downstream SR quality (lower CLIP-IQA+ and MUSIQ), while the causal memory network produces spatiotemporally coherent latents that substantially ease the SR task. Increasing the number of memory blocks from 4 to 8 further improves quality with marginal overhead ($<$5\% of total pipeline latency).

\noindent\textbf{Cascaded Streaming Optimization.}
We ablate each component of the streaming optimization scheme in Table~\ref{tab:ablation_opt}.
\textbf{\emph{(i) Sparse attention:}} Dense causal attention triggers out-of-memory at $960{\times}1664$; block-sparse attention resolves this. Content-adaptive top-$k$ selection (vs.\ fixed window) yields better quality by allowing global information routing where needed.
\textbf{\emph{(ii) Hybrid reward:}} Adding CLIP-IQA$^+$, MUSIQ, and LAION-Aesthetic reward signals during distillation improves perceptual quality beyond what Decoupled DMD + reconstruction losses alone achieve.
\textbf{\emph{(iii) DPO preference optimization:}} The cascaded DPO loss with a stronger Wan2.2-5B teacher closes the train-test gap, improving temporal consistency and visual quality over long streaming sequences.
\textbf{\emph{(iv) Dynamic cache management:}} The three-pronged inference optimization (step reduction, IQA-adaptive cache refresh, SR cache length adaptation) significantly improves FPS with negligible quality impact.

\subsection{Qualitative Results}
Figure~\ref{fig:qualitative2}, \ref{fig:qualitative3} and \ref{fig:qualitative4} presents visual comparisons. Prior streaming methods generate at $480{\times}832$ with visible artifacts (blurriness, flickering), while \method{} produces sharp, temporally coherent frames at 1K \& 2K with rich high-frequency detail. The causal streaming latent upsampler avoids aliasing artifacts visible in naive-interpolation baselines, the hybrid reward signals yield sharper textures and more natural colors, and the cascaded DPO preserves quality over extended sequences.

\section{Conclusion}
\label{sec:conclusion}
We presented \method{}, a cascaded streaming framework that scales real-time autoregressive video generation from 480P to 1K (${\sim}$30 FPS) and 2K (${\sim}$18 FPS) on a single GPU. The key contributions---architecture-preserving SR training with AIGC degradation, a causal streaming latent upsampler with HR decoder, and cascaded streaming optimization via sparse distillation and DPO---jointly advance the resolution--speed Pareto frontier while maintaining state-of-the-art visual quality. Limitations and broader impact are discussed in Appendix~\ref{app:discussion}.

\newpage
\bibliographystyle{iclr2026_conference}
\bibliography{ref}

\newpage
\appendix

\section{Related Work}
\label{app:related}

\paragraph{Video Diffusion Models.}
Recent advances in video generation have achieved remarkable progress, enabling the synthesis of high-fidelity and temporally coherent videos directly from textual prompts~\cite{yang2024cogvideox, kong2024hunyuanvideo, wan2025wan}. Early methods extend text-to-image diffusion models by introducing temporal modules to capture frame dynamics, yet often fail to model holistic spatiotemporal dependencies~\cite{ho2022video,guo2023animatediff,blattmann2023stable}. With the emergence of the diffusion transformer (DiT)~\cite{peebles2023scalable}, transformer-based architectures have become the dominant paradigm, jointly modeling spatial and temporal correlations through full attention mechanisms~\cite{yang2024cogvideox, kong2024hunyuanvideo,hacohen2024ltx,wan2025wan}. Modern text-to-video (T2V) models typically adopt a framework consisting of a 3D VAE for spatiotemporal compression and a DiT for latent-space denoising. Building on this foundation, recent works, including CogVideoX~\cite{yang2024cogvideox}, HunyuanVideo~\cite{kong2024hunyuanvideo}, and Wan~\cite{wan2025wan}, further scale up model size and data, demonstrating impressive video quality at unprecedented levels. However, their non-autoregressive, bidirectional attention structure prevents streaming and incurs high latency for interactive use.

\paragraph{Autoregressive Video Diffusion.}
Given the inherent temporal order of video data, it is natural to model video generation as an autoregressive process. While most video diffusion models rely on bidirectional dependencies~\cite{videoworldsimulators2024,yang2024cogvideox,polyak2024movie}, autoregressive video diffusion has recently been explored under two paradigms. \emph{Teacher Forcing}~\cite{valevski2024diffusion,jin2024pyramidal} trains models to denoise new frames conditioned on clean context frames, while \emph{Diffusion Forcing}~\cite{chen2024diffusion,ruhe2024rolling,kim2024fifo} supports autoregressive sampling by conditioning on frames with varying noise levels. CausVid~\cite{yin2025causvid} first adapted bidirectional DiTs to autoregressive generation with causal attention, using ODE-trajectory initialization and asymmetric distribution matching distillation (DMD)~\cite{yin2024dmd} to reduce denoising steps. Self Forcing~\cite{huang2025selfforcing} addressed the critical issue of \emph{exposure bias}---the train-test mismatch where models trained on ground-truth context must generate conditioned on their own imperfect outputs---by simulating autoregressive rollouts during training. Causal Forcing~\cite{zhu2026causal} improved upon both with better ODE initialization and causal consistency distillation. More recent works extend this direction: Self-Forcing++~\cite{cui2025selfforcingpp} scales to minute-length videos, Reward Forcing~\cite{lu2025rewardforcing} integrates reward signals into streaming distillation, and DummyForcing~\cite{guo2026dummy} exploits redundant attention heads for training-free acceleration. Despite these advances, all existing methods remain confined to low resolutions ($480$P) and none addresses the quadratic attention bottleneck that prohibits high-resolution streaming.

\paragraph{Ultra-High-Resolution Video Generation.}
Ultra-high-resolution (UHR) video generation remains a fundamental challenge, hindered by immense computational demands and the scalability constraints of current models~\cite{xue2025ultravideo,lu2026omnivr}. Existing research primarily follows three paradigms. \emph{Training-free methods} extend pre-trained diffusion models to higher resolutions without retraining by modifying denoising processes or attention structures~\cite{he2023scalecrafter, du2024demofusion}, achieving computational efficiency but often producing over-smoothed textures and lacking genuine high-frequency detail. \emph{Fine-tuning strategies} adapt low-resolution generative models on high-resolution datasets~\cite{cheng2024resadapter, ren2024ultrapixel}, enhancing fidelity while preserving generative priors. \emph{Cascaded methods} have recently emerged as a promising direction: FlashVideo~\cite{zhang2025flashvideo} and LongCat~\cite{team2025longcat} adopt two-stage pipelines with low-resolution generation followed by high-resolution refinement; Seedance~\cite{gao2025seedance} demonstrates that motion dynamics are more effectively learned at lower resolutions; HunyuanVideo~1.5~\cite{wu2025hunyuanvideo15} and FSVideo~\cite{fsvideo2026} employ latent-space upsampling followed by a large SR model but with no streaming capability; DaVinci~\cite{sii2026davinci} directly applies interpolation upsampling in latent space, introducing frequency aliasing that burdens the subsequent SR stage; LTX-2~\cite{hacohen2026ltx2} introduces a latent upsampler for multi-scale generation, but its upsampler is heavyweight and sensitive to sequence length, making it unsuitable for arbitrary-length streaming. These cascaded approaches are not designed for causal streaming and typically require heavy noise injection to compensate for upsampling artifacts, increasing SR training difficulty. Our work addresses these limitations by performing latent-space upsampling with a spatiotemporally coherent causal memory network, enabling real-time high-resolution streaming within a unified cascaded pipeline.

\paragraph{Efficient Attention for Video.}
The quadratic cost of attention is a primary bottleneck for high-resolution video. Sparse attention patterns---including local windows~\cite{beltagy2020longformer}, strided patterns~\cite{child2019generating}, and learned sparsity~\cite{kitaev2020reformer}---have been extensively explored for language and images. For video, SpargeAttn~\cite{zhang2025spargeattn} and Block-Sparse Attention~\cite{blocksparse2025} provide hardware-efficient sparse kernels that accelerate attention computation. FlashVSR~\cite{zhuang2025flashvsr} was the first to apply block-sparse attention to video super-resolution, introducing locality-constrained sparse patterns and a progressive distillation pipeline from dense to sparse attention. DummyForcing~\cite{guo2026dummy} observed that ${\sim}25$\% of attention heads in autoregressive video DiTs are ``dummy'' (attending only to the current frame) and exploited this for inference-time acceleration. However, FlashVSR requires projecting pixel-space low-quality video into the DiT latent space for conditional super-resolution, while DummyForcing uses sparsity only at inference without training adaptation. In contrast, our work integrates dynamic block-sparse attention into a pure generative streaming pipeline without any pixel-space input dependency, and achieves higher sparse efficiency through content-adaptive mask prediction within causal autoregressive distillation.

\paragraph{Distillation for Fast Generation.}
Distribution matching distillation (DMD)~\cite{yin2024dmd,yin2024dmd2} reduces multi-step diffusion to few-step generation by matching output distributions via a learned critic. Decoupled DMD~\cite{liu2025decoupledDMD} further improves this paradigm by separating CFG augmentation from distribution matching, achieving better quality--speed trade-offs. Consistency distillation~\cite{song2023consistency} and rectified flow~\cite{liu2022flow} offer alternative paths to fast sampling. In the video domain, CausVid~\cite{yin2025causvid} and Causal Forcing~\cite{zhu2026causal} extend DMD to autoregressive video with asymmetric teacher--student training, but use dense attention throughout. Reward Forcing~\cite{lu2025rewardforcing} incorporates reward signals into DMD for streaming video but remains at low resolution. We build upon Decoupled DMD and augment it with perceptual and aesthetic reward signals~\cite{xu2024refl,ke2021musiq}, directly optimizing for perceptual quality rather than surrogate losses, while simultaneously training with sparse causal attention to enable single-step high-resolution streaming generation.

\paragraph{Video Super-Resolution.}
Diffusion-based SR models~\cite{zhuang2025flashvsr,he2024venhancer,lu2026expert,Lu2025Tone,Lu2025Deblurring,fan2026bird,lu2025elucidating,Lu_2024_CVPR_HirFormer,Lu_2025_CVPR_AALN,Lu_2025_CVPR_EvenFormer,Lu_2025_CVPR_ILAWR,wang2025ntire,peng2020cumulative,wang2023decoupling,peng2024towards,peng2021ensemble,ren2024ultrapixel,peng2024efficient,peng2025pixel,peng2024unveiling,pengboosting} achieves high fidelity but is computationally heavy, often requiring multiple denoising steps per frame. FlashVSR~\cite{zhuang2025flashvsr} significantly accelerated diffusion-based VSR through block-sparse causal attention and a tiny conditional decoder, achieving near real-time streaming at $768{\times}1408$. Stream-DiffVSR~\cite{shiu2025streamdiffvsr} further explored autoregressive causal conditioning for low-latency streaming SR. However, both methods fundamentally operate as conditional super-resolution models that project pixel-space low-quality frames into the DiT latent space, requiring explicit LQ video input and architectural modifications that forfeit the generative capability of the pre-trained T2V model. Moreover, as Waver~\cite{zhang2025waver} demonstrates, composing pixel-space SR with latent-space generators introduces additional encode--decode overhead that limits end-to-end efficiency. In contrast, our framework is a pure generative streaming pipeline that produces high-resolution video directly from text via an architecture-preserving training paradigm with an AIGC-oriented degradation pipeline, preserving the base model's generative priors. The streaming latent upsampler performs resolution scaling entirely in latent space with a causal memory network, seamlessly integrating into the end-to-end cascaded streaming pipeline without any pixel-space dependency.

\section{Limitations and Broader Impact}
\label{app:discussion}

\noindent\textbf{Limitations.}
(1)~The block-sparse attention kernel achieves optimal hardware utilization on H200/B200 GPUs; performance on consumer-grade GPUs remains less optimized.
(2)~The DPO positive samples rely on a stronger Wan2.2-5B model; exploring online reward-based preference optimization could eliminate this dependency.
(3)~The current framework achieves real-time streaming at up to 2K resolution; scaling to 4K real-time generation remains beyond reach due to the quadratic growth of attention cost and memory bandwidth constraints. Achieving 4K real-time streaming video generation is a primary future research direction, potentially requiring advances in sub-linear attention mechanisms, more aggressive model compression, and hardware-software co-design.

\noindent\textbf{Broader Impact.}
Real-time high-resolution video generation has broad positive potential in creative industries, accessibility, education, and interactive media. We acknowledge risks associated with deepfakes and misinformation, and advocate for robust watermarking, content provenance tracking, and responsible deployment practices.

\section{Implementation Details}
\label{app:implementation}

\noindent\textbf{Training Protocol.}
Training proceeds in four stages:
\emph{(i)~Architecture-preserving SR fine-tuning} (\S\ref{sec:sr_training}): the T2V model is converted to a TV2V SR model via zero-initialized channel extension and trained on AIGC-degraded data with the flow matching objective (Eq.\,\ref{eq:reg}), condition noise injection ($\sigma_{\text{cond}} \in [0.4, 0.6]$), and condition dropout ($p_{\text{drop}}{=}0.4$).
\emph{(ii)~Causal memory network pre-training} (\S\ref{sec:upsampler}): the streaming latent upsampler and HR decoder are trained on paired low-/high-resolution data with latent MSE and eWarp temporal consistency losses (Eq.\,\ref{eq:upsampler_loss}).
\emph{(iii)~Hybrid-reward-enhanced sparse causalization and single-step distillation} (Phase~I, \S\ref{sec:sparse_distill}): the SR model is distilled via Decoupled DMD~\cite{liu2025decoupledDMD} with block-sparse causal attention (block size $(2,8,8)$, adaptive top-$k$, local window $9{\times}9$), wavelet L1 + LPIPS reconstruction losses, and hybrid reward signals (CLIP-IQA$^+$, MUSIQ, LAION-Aesthetic) at high resolution ($960{\times}1664$).
\emph{(iv)~Cascaded streaming DPO} (Phase~II, \S\ref{sec:self_forcing}): the full cascaded pipeline is jointly rolled out, and a DPO loss (Eq.\,\ref{eq:pref}) aligns the SR model's output with a stronger Wan2.2-5B model.
All stages use 32 GPUs (4 nodes $\times$ 8 H200), AdamW ($\text{lr}{=}10^{-5}$, $\beta{=}(0.9, 0.95)$), gradient clipping at 1.0, and bf16 mixed precision.

\noindent\textbf{SR DiT Architecture.}
The SR model follows the Wan2.1 architecture with a $2c$-channel input (16 noise + 16 condition latent, zero-initialized extension):
30 transformer blocks, hidden dim 1536, FFN dim 8960, 12 heads (head dim 128);
patch size $(1, 2, 2)$; 3D RoPE with axis dims $(44, 42, 42)$;
sparse block size $(2, 8, 8)$; adaptive top-$k$ with $\rho{=}1.0$, $S_{\text{ref}}{=}1560$; local window $9{\times}9$;
streaming chunk size: 2 latent frames (8 pixel frames); KV cache: 3 temporal windows.

\noindent\textbf{Causal Streaming Latent Upsampler.}
Three-stage architecture inspired by TAEHV~\cite{ollin2024taehv}: stage channels $[256, 128, 64]$, 3 CausalMemBlocks per stage; spatial factors $[2, 1, 1]$ (PixelShuffle~\cite{shi2016pixelshuffle} for $2{\times}$ spatial upsampling in stage 1); temporal factors $[1, 1, 1]$; 16 input/output channels. Total: ${\sim}$2.1M parameters. Trained with latent MSE + eWarp loss, lr$=2{\times}10^{-4}$, cosine scheduler.

\noindent\textbf{HR Decoder.}
Same CausalMemBlock architecture: stage channels $[256, 128, 64, 64]$; spatial factors $[2, 2, 2]$ (PixelShuffle); temporal factors $[1, 2, 2]$; 16 latent input channels, 3 RGB output channels. Supports parallel (training) and sequential (streaming inference) execution modes.

\noindent\textbf{Training Hyperparameters.}
\emph{SR fine-tuning}: flow matching with log-normal sigma sampling (flow shift $s{=}1.5$); condition noise $\sigma_{\text{cond}} \sim \mathcal{U}[0.4, 0.6]$; condition dropout $p_{\text{drop}}{=}0.4$; CFG rate 0.1; AdamW, $\text{lr}{=}10^{-5}$, $\beta_1{=}0.9$, $\beta_2{=}0.95$; gradient clipping 10.0; bf16 precision.
\emph{Phase~I distillation}: Decoupled DMD with CA schedule $\tau_{\text{CA}} > t$ and DM schedule $\tau_{\text{DM}} \in [0,1]$; fake score update ratio 5$\times$; teacher: 20-step inference, guidance scale 3.5; reconstruction: wavelet L1 (HF sub-bands) + LPIPS; hybrid reward: CLIP-IQA$^+$, MUSIQ, LAION-Aesthetic.
\emph{Phase~II DPO}: positive samples from Wan2.2-5B multi-step pixel-space SR; temperature $\beta{=}0.1$; reference policy: frozen Phase~I checkpoint.
EMA: decay 0.99, start step 3000, update every 5 steps.

\noindent\textbf{AIGC Degradation Pipeline.}
Stage~1 (AIGC synthetic): temporal morphing ($\alpha \in [0.2, 0.9]$), stochastic frame drop + interpolation, directional motion blur (smooth angle/length trajectories), ROI-constrained grid warping (max displacement 14px), H.264 FFmpeg compression (CRF $[25, 30]$).
Stage~2 (Real-ESRGAN-style): two passes of USM sharpening $\rightarrow$ Gaussian blur (kernel $[15,37]$, $\sigma \in [0.2, 3.0]$) $\rightarrow$ random rescaling ($[0.15, 1.5]$) $\rightarrow$ additive noise $\rightarrow$ JPEG ($q \in [70, 95]$).
Final: bicubic $2{\times}$--$4{\times}$ downsampling + upsample; stochastic CutMix mixing between AIGC and spatial branches.

\noindent\textbf{Dynamic Cache Management (Inference).}
LR generator step reduction: 4 steps for the first chunk, 3 steps for subsequent chunks.
IQA-adaptive cache refresh: skip $\mathbf{x}_0$ KV forward when previous chunk IQA exceeds threshold.
SR cache window: dynamically selected based on content complexity.

\section{Additional Qualitative Results}
\label{app:qualitative}

We present additional qualitative comparisons to further demonstrate the visual quality and generalization capability of \method{} across diverse scenes, subjects, and motion patterns. All results are generated in a fully streaming fashion on a single GPU, with \method{} operating at real-time throughput (${\sim}30$ FPS at 1K, ${\sim}18$ FPS at 2K). For each example, we compare frames generated by prior 480P streaming methods (Self Forcing~\cite{huang2025selfforcing}, CausVid~\cite{yin2025causvid}, DummyForcing~\cite{guo2026dummy}) against the high-resolution output of \method{} at $960{\times}1664$ and $1440{\times}2560$. Zoomed-in crops highlight fine-grained differences in texture fidelity, temporal coherence, and aesthetic quality.

\begin{figure}[ht]
  \centering
  \includegraphics[width=\textwidth]{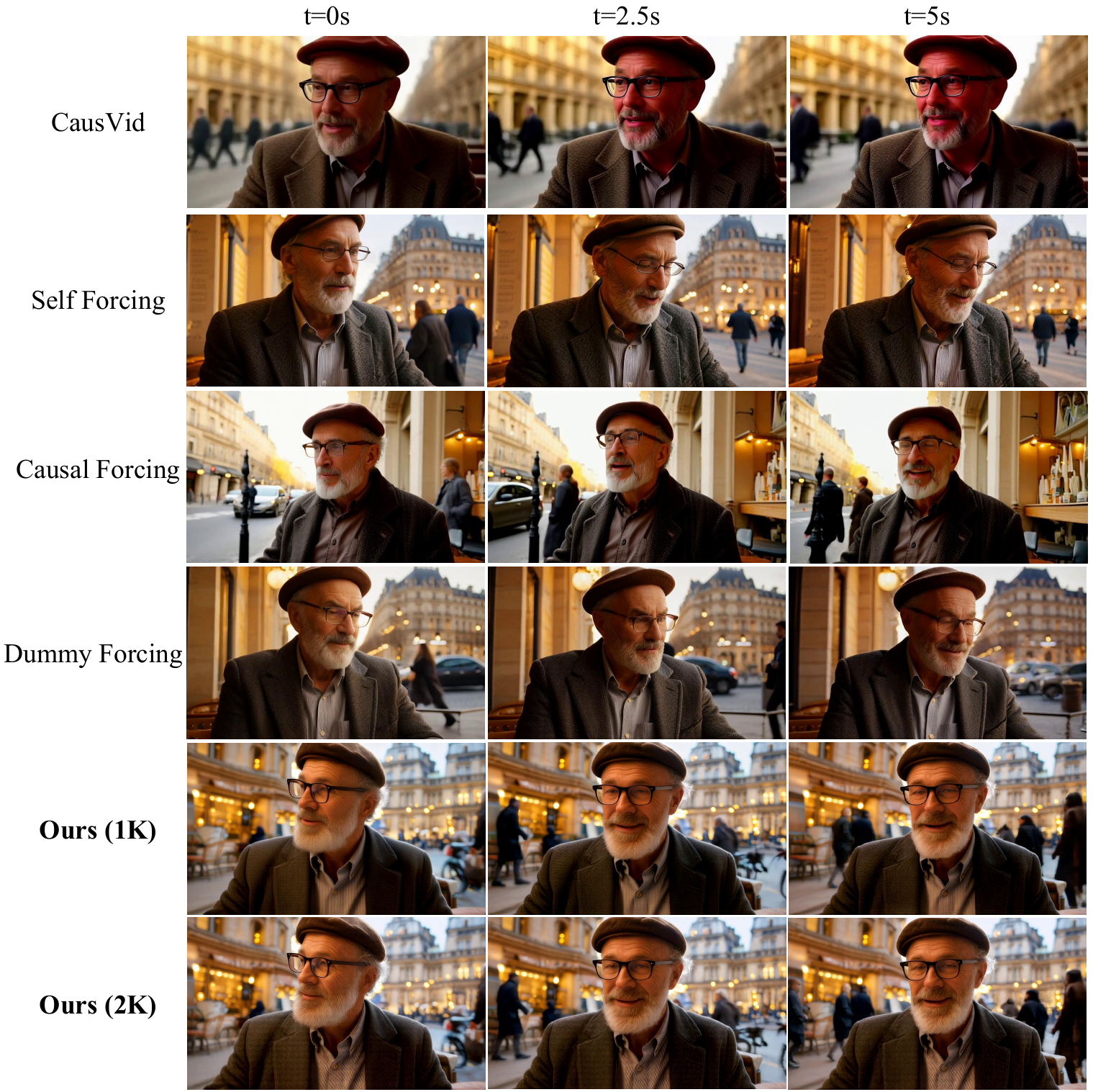}
  \caption{\textbf{Fine-grained texture comparison.} \method{} resolves high-frequency details---individual hair strands, skin texture, fabric patterns---that are lost in 480P baselines. Zoomed crops (bottom) highlight the substantial resolution advantage of our pipeline.}
  \label{fig:compare2}
\end{figure}

\paragraph{Fine-Grained Texture Fidelity.}
As shown in Fig.~\ref{fig:compare2}, \method{} exhibits substantially superior texture detail compared to 480P baselines. In close-up portrait scenes, individual strands of hair, pore-level skin texture, and fine fabric weaves are clearly resolved at 1K and 2K resolution, whereas baseline methods produce visibly blurred or over-smoothed results. This improvement stems from the combination of the AIGC-oriented degradation pipeline---which trains the SR model to faithfully restore AI-generated textures---and the causal streaming latent upsampler, which provides spatiotemporally coherent latent inputs free of frequency aliasing.

\begin{figure}[ht]
  \centering
  \includegraphics[width=\textwidth]{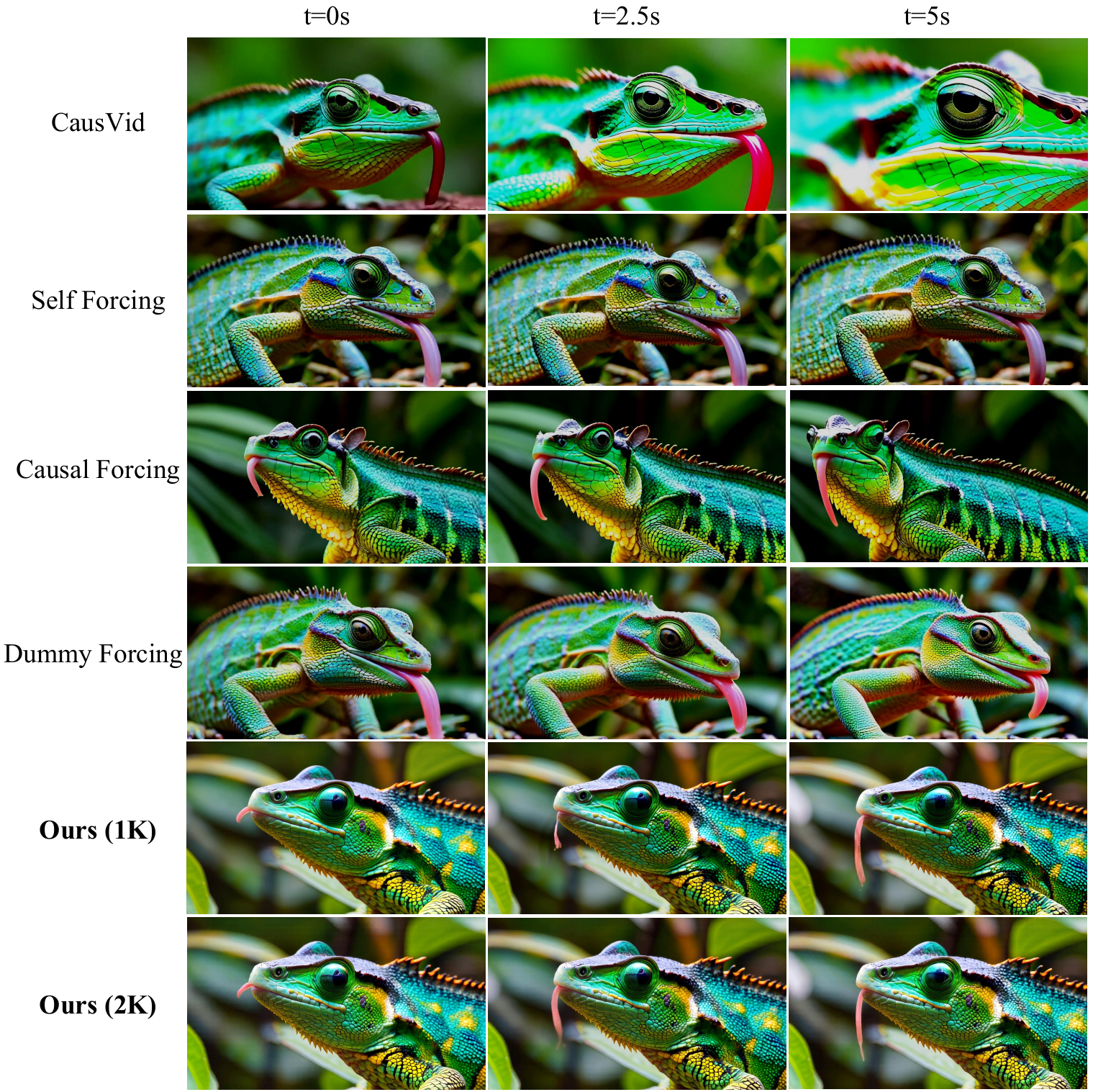}
  \caption{\textbf{Temporal consistency and color stability.} On scenes with intricate surface details (\eg, animal skin, vegetation), \method{} maintains consistent exposure, stable color, and temporally coherent textures, while baselines exhibit color drift and flickering artifacts.}
  \label{fig:compare5}
\end{figure}

\paragraph{Temporal Consistency and Color Stability.}
Fig.~\ref{fig:compare5} demonstrates the temporal consistency of \method{} on challenging natural scenes with complex surface details. Prior methods exhibit noticeable color drift, exposure fluctuation, and temporal flickering across frames, especially on high-frequency surfaces such as animal skin and intricate vegetation. In contrast, \method{} maintains stable exposure, vivid and consistent color reproduction, and temporally coherent details throughout the sequence. This robustness is attributed to the cascaded DPO with the stronger Wan2.2-5B teacher, which explicitly optimizes for temporal coherence over extended streaming rollouts.

\begin{figure}[ht]
  \centering
  \includegraphics[width=\textwidth]{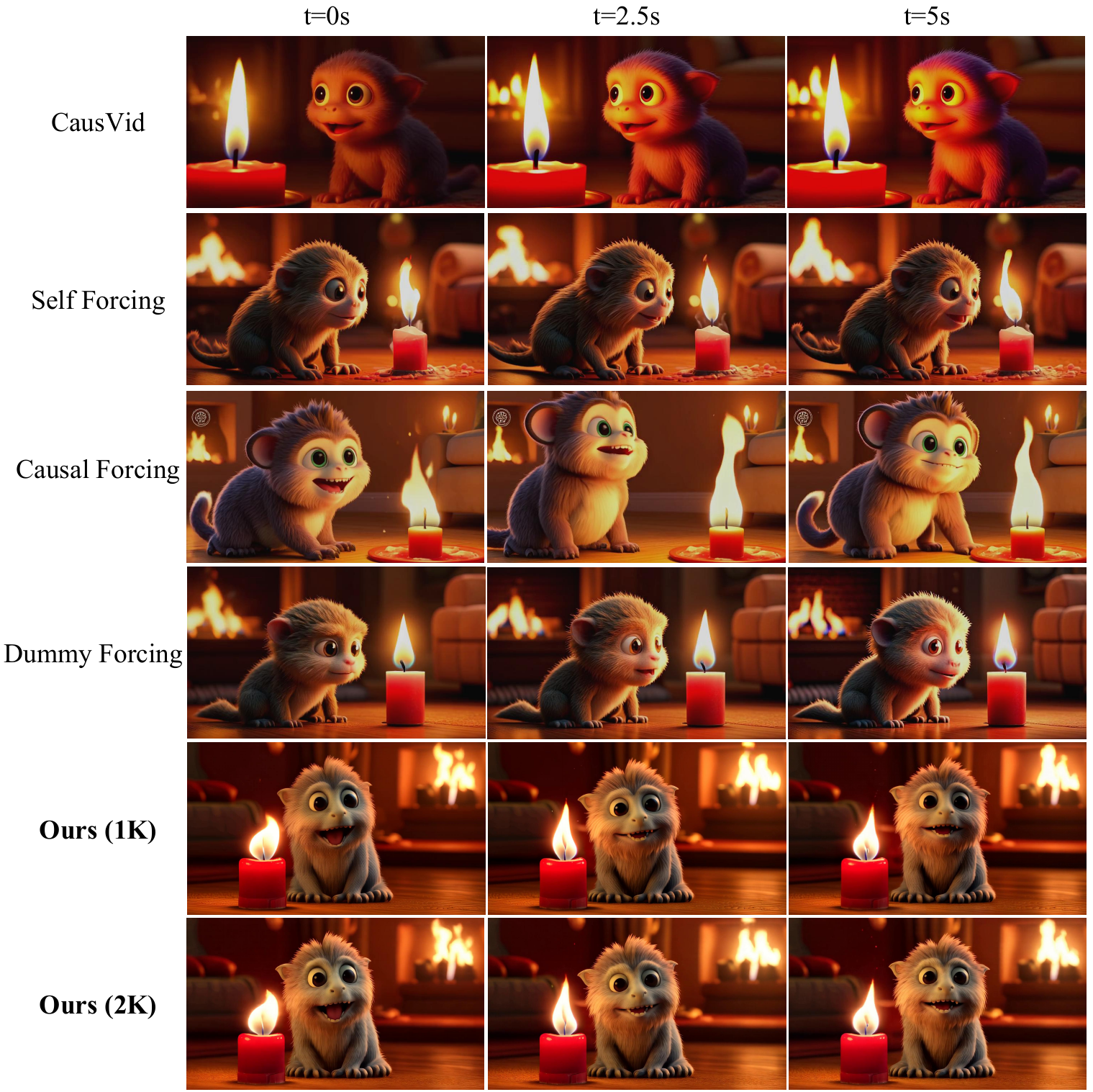}
  \caption{\textbf{Complex scene composition.} \method{} accurately renders multi-object scenes with rich spatial structure, maintaining sharp foreground details and coherent backgrounds at high resolution. Baselines produce muddled textures and lose fine structural detail.}
  \label{fig:compare3}
\end{figure}

\paragraph{Complex Scene Composition.}
Fig.~\ref{fig:compare3} presents comparisons on scenes with complex spatial compositions involving multiple objects, varied depths, and rich background detail. \method{} faithfully renders both foreground subjects and background elements at high resolution, preserving sharp edges, clear object boundaries, and natural depth-of-field effects. Baseline methods, constrained to 480P, struggle to separate fine foreground details from the background, often producing muddled textures and lost structural detail in peripheral regions.

\begin{figure}[ht]
  \centering
  \includegraphics[width=\textwidth]{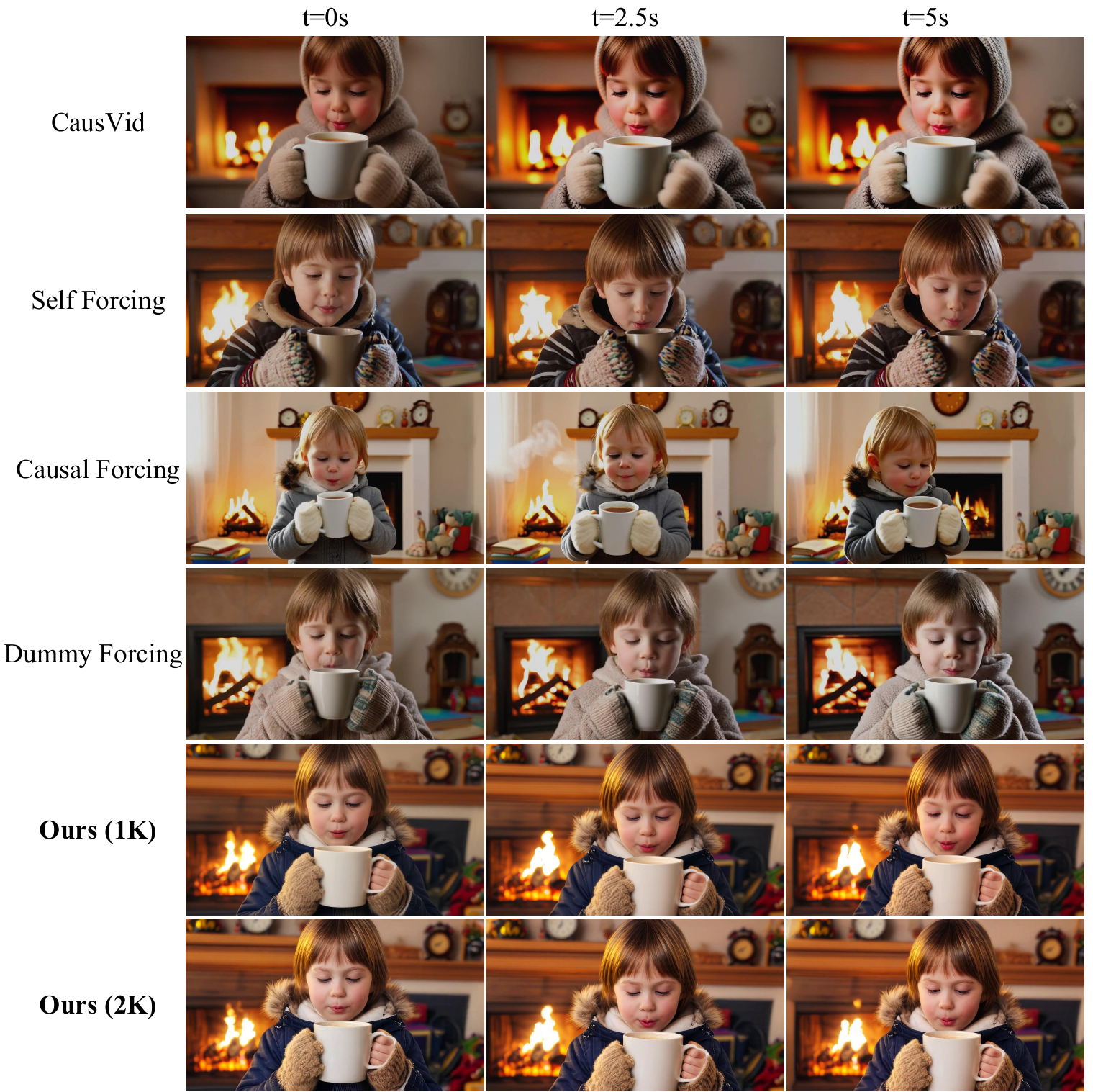}
  \caption{\textbf{Dynamic motion and semantic coherence.} Under fast camera movements and complex object interactions, \method{} produces temporally smooth, artifact-free high-resolution frames, while baselines exhibit motion blur, ghosting, and temporal inconsistencies.}
  \label{fig:compare4}
\end{figure}

\paragraph{Dynamic Motion and Semantic Coherence.}
Fig.~\ref{fig:compare4} showcases scenes with significant dynamic motion, including fast camera movements, object interactions, and complex temporal dynamics. \method{} generates temporally smooth and semantically coherent high-resolution frames even under rapid motion, without introducing motion blur artifacts, ghosting, or temporal discontinuities. The hybrid reward signals (CLIP-IQA$^+$, MUSIQ, LAION-Aesthetic) during distillation ensure that perceptual quality is preserved under dynamic conditions, while the dynamic cache management strategy maintains generation efficiency without sacrificing quality during fast-paced sequences.

\section{VBench Scores Across All Dimensions}
\label{app:vbench}

To provide a comprehensive evaluation beyond the aggregate metrics reported in the main paper, we evaluate \method{} on all 16 VBench~\cite{huang2024vbench} dimensions and compare against representative methods: the Wan2.1-1.3B teacher model (50 steps), CausVid~\cite{yin2025causvid} (4 steps), Self Forcing~\cite{huang2025selfforcing} (4 steps), and DummyForcing~\cite{guo2026dummy} (4 steps). All methods use the same base architecture (Wan2.1-1.3B) and evaluation prompts with prompt rewriting via Qwen2.5-7B-Instruct.

\begin{figure}[ht]
  \centering
  \includegraphics[width=0.75\textwidth]{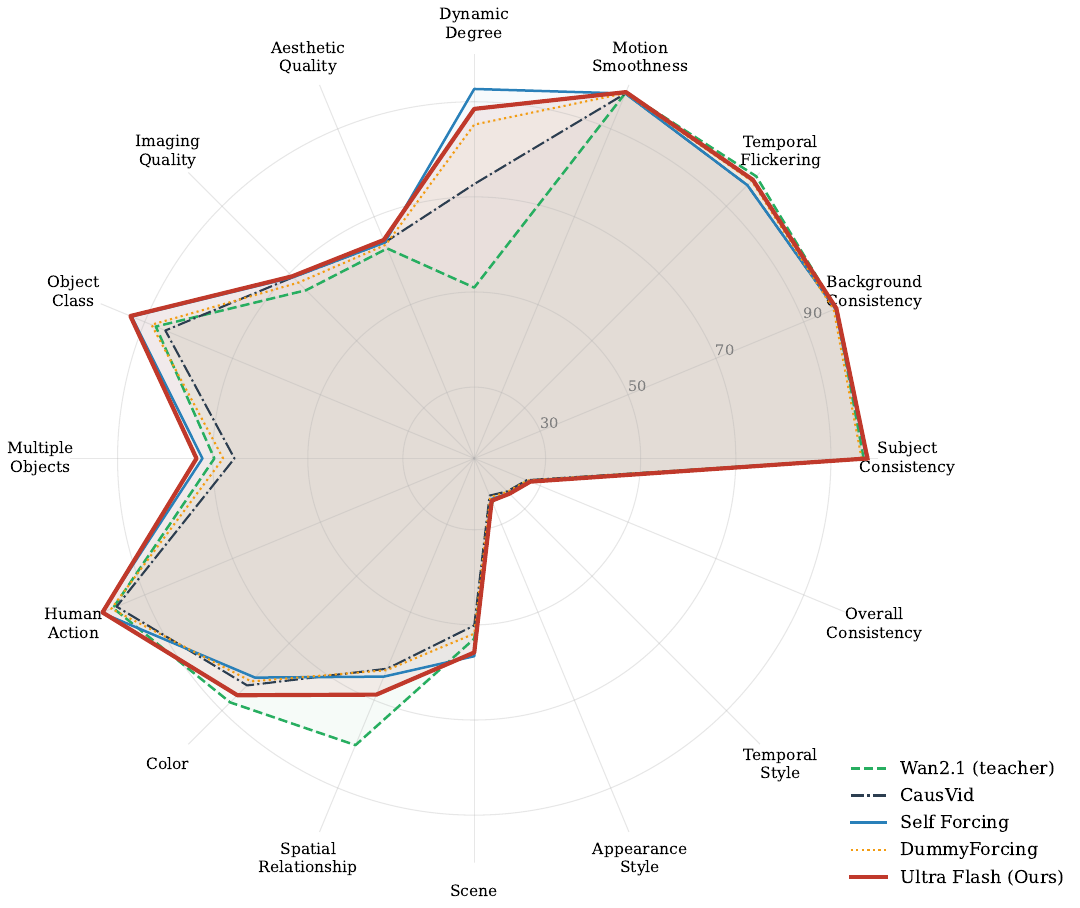}
  \caption{\textbf{VBench 16-dimension radar chart.} We compare \method{} with Wan2.1 (teacher), CausVid, Self Forcing, and DummyForcing across all 16 VBench metrics. \method{} achieves the best or near-best performance across most dimensions while maintaining real-time throughput.}
  \label{fig:vbench_radar}
\end{figure}

As shown in Fig.~\ref{fig:vbench_radar}, \method{} achieves the best or competitive performance across the majority of the 16 VBench dimensions. Several observations are worth noting:

\paragraph{Temporal Quality.} \method{} achieves the highest temporal flickering score (97.85) and motion smoothness (98.37) among all single-step methods, surpassing even the 50-step Wan2.1 teacher in temporal flickering. This demonstrates that our cascaded self-forcing preference optimization effectively maintains temporal coherence during high-resolution streaming generation, even when operating in single-step mode.

\paragraph{Frame Quality.} \method{} achieves the best imaging quality (69.15) and aesthetic quality (64.72), outperforming all baselines including the multi-step teacher. The hybrid reward integration (CLIP-IQA$^+$, MUSIQ, LAION-Aesthetic) during Phase~I distillation directly optimizes these perceptual quality metrics, while the high-resolution generation further enhances fine-grained visual quality.

\paragraph{Semantic Alignment.} In semantic dimensions---object class (93.25), multiple objects (73.40), and human action (99.60)---\method{} achieves strong scores competitive with Self Forcing. The architecture-preserving training paradigm ensures that the original T2V model's semantic understanding is retained through the SR conversion process.

\paragraph{Dynamic Degree.} \method{} achieves a dynamic degree of 88.50, which is slightly lower than Self Forcing's 92.69 but significantly higher than Wan2.1 (50.93) and CausVid (72.69). This indicates that our cascaded pipeline preserves dynamic motion well despite the additional SR processing, and the preference optimization prevents the model from collapsing to static outputs.

\paragraph{Style and Consistency.} In appearance style, temporal style, and overall consistency, all methods perform comparably since these dimensions are largely determined by the base model's pre-trained knowledge. \method{} achieves marginally higher scores (24.62, 25.48, 27.75) due to the enhanced visual quality from high-resolution generation.

\section{Detailed Algorithm Descriptions}
\label{app:algorithms}

We provide detailed pseudocode for the three core algorithmic contributions of \method{}: the architecture-preserving SR training paradigm (Algorithm~\ref{alg:sr_training}), the causal streaming latent upsampler inference (Algorithm~\ref{alg:upsampler}), and the cascaded streaming optimization and inference pipeline (Algorithm~\ref{alg:cascaded_opt}).

\begin{algorithm}[ht]
\caption{Architecture-Preserving T2V-to-TV2V SR Training}
\label{alg:sr_training}
\footnotesize
\begin{algorithmic}[1]
\REQUIRE Pre-trained T2V model $f_\theta$ (Wan2.1-1.3B), HR video dataset $\mathcal{D}$, AIGC degradation pipeline $\mathcal{A}$
\ENSURE Trained multi-step SR model $f_\theta^{\text{SR}}$
\STATE Extend input projection of $f_\theta$: channels $c \to 2c$, zero-init new weights
\FOR{each training iteration}
  \STATE Sample HR video clip $\mathbf{I}_{\text{gt}} \in \mathcal{D}$, encode to latent $\mathbf{z}_0 = \mathcal{E}(\mathbf{I}_{\text{gt}})$
  \STATE \textcolor{blue}{\textit{// AIGC-Oriented Degradation Pipeline}}
  \STATE Apply Stage-1 AIGC degradation: temporal morphing, frame drop, motion blur, grid warp, codec
  \STATE Apply Stage-2 spatial degradation: USM $\to$ blur $\to$ resize $\to$ noise $\to$ JPEG ($\times 2$ passes)
  \STATE Apply $2{\times}$--$4{\times}$ bicubic downsampling then upsample back
  \STATE With prob $p_{\text{mix}}$: CutMix AIGC-degraded and spatial-degraded branches
  \STATE Obtain degraded latent $\mathbf{z}^{\text{LR}}$
  \STATE \textcolor{blue}{\textit{// Latent Upsampling (simulated during training)}}
  \STATE $\mathbf{z}^{\text{HR}} \leftarrow \text{Upsample}(\mathbf{z}^{\text{LR}})$ \hfill\textit{(via causal memory network or bicubic)}
  \STATE \textcolor{blue}{\textit{// Conditioning Augmentation}}
  \STATE Sample $\sigma_{\text{cond}} \sim \mathcal{U}[\sigma_{\min}, \sigma_{\max}]$; add noise: $\mathbf{z}^{\text{HR}} \leftarrow \mathbf{z}^{\text{HR}} + \sigma_{\text{cond}} \cdot \boldsymbol{\epsilon}'$
  \STATE With prob $p_{\text{drop}}$: set $\mathbf{z}^{\text{HR}} \leftarrow \mathbf{0}$ \hfill\textit{(condition dropout for CFG)}
  \STATE \textcolor{blue}{\textit{// Flow Matching Training}}
  \STATE Sample timestep $t \sim p(t)$, noise $\boldsymbol{\epsilon} \sim \mathcal{N}(0, \mathbf{I})$
  \STATE Construct $\mathbf{z}_t = (1 - \sigma_t)\mathbf{z}_0 + \sigma_t \boldsymbol{\epsilon}$
  \STATE Concatenate input: $\mathbf{x}_{\text{in}} = [\mathbf{z}_t\,;\, \mathbf{z}^{\text{HR}}]$ \hfill\textit{(channel dim $2c$)}
  \STATE Compute loss: $\mathcal{L}_{\text{FM}} = \| f_\theta(\mathbf{x}_{\text{in}}, t, \mathbf{c}_{\text{text}}) - (\boldsymbol{\epsilon} - \mathbf{z}_0) \|_2^2$
  \STATE Update $\theta$ via gradient descent on $\mathcal{L}_{\text{FM}}$
\ENDFOR
\RETURN $f_\theta^{\text{SR}}$
\end{algorithmic}
\end{algorithm}

\begin{algorithm}[ht]
\caption{Causal Streaming Latent Upsampler \& HR Decoder}
\label{alg:upsampler}
\footnotesize
\begin{algorithmic}[1]
\REQUIRE LR latent sequence $\{\mathbf{z}^{\text{LR}}_1, \ldots, \mathbf{z}^{\text{LR}}_T\}$, trained causal memory network $\mathcal{U}$
\REQUIRE Config: spatial\_factors $[r_1, r_2, r_3]$, temporal\_factors $[s_1, s_2, s_3]$, $N_b$ blocks/stage
\ENSURE HR latent/pixel sequence $\{\hat{\mathbf{z}}^{\text{HR}}_1, \ldots, \hat{\mathbf{z}}^{\text{HR}}_T\}$
\STATE
\STATE \textcolor{blue}{\textbf{--- Variant A: Parallel Inference (Training) ---}}
\STATE \textcolor{blue}{\textit{// All $T$ frames processed simultaneously; causal memory via temporal shift}}
\STATE $\mathbf{H} \leftarrow \text{Conv}_{3\times3}^{\text{in}}(\mathbf{Z}^{\text{LR}})$ \hfill\textit{($\mathbf{Z}^{\text{LR}} \in \mathbb{R}^{T \times C \times H \times W}$)}
\FOR{stage $s = 1, 2, 3$}
  \FOR{block $b = 1, \ldots, N_b$}
    \STATE $\mathbf{M} \leftarrow \text{TemporalShift}(\mathbf{H})$ \hfill\textit{($\mathbf{M}_t = \mathbf{H}_{t-1}$, $\mathbf{M}_1 = \mathbf{0}$)}
    \STATE $\mathbf{H}_{\text{cat}} \leftarrow [\mathbf{H}\,;\, \mathbf{M}]$ \hfill\textit{(channel concat, all frames in parallel)}
    \STATE $\mathbf{H}_{\text{conv}} \leftarrow \text{Conv}^{(3)}(\text{Conv}^{(2)}(\text{ReLU}(\text{Conv}^{(1)}(\mathbf{H}_{\text{cat}}))))$
    \STATE $\mathbf{H} \leftarrow \text{ReLU}(\mathbf{H}_{\text{conv}} + W_{\text{skip}} \cdot \mathbf{H})$
  \ENDFOR
  \STATE $\mathbf{H} \leftarrow \text{PixelShuffle}_{r_s}(\text{Conv}_{3\times3}^{\text{up}_s}(\mathbf{H}))$
  \IF{$s_s > 1$}
    \STATE $\mathbf{H} \leftarrow \text{Reshape}(\text{Conv}_{1\times1}^{\text{texp}_s}(\mathbf{H}))$ \hfill\textit{($T \to T \cdot s_s$)}
  \ENDIF
  \STATE $\mathbf{H} \leftarrow \text{Conv}_{1\times1}^{\text{ch}_s}(\mathbf{H})$
\ENDFOR
\STATE $\hat{\mathbf{Z}}^{\text{HR}} \leftarrow \text{Conv}_{3\times3}^{\text{out}}(\mathbf{H})$
\RETURN $\hat{\mathbf{Z}}^{\text{HR}}$ \hfill\textit{(all frames, supports gradient back-propagation)}
\STATE
\STATE \textcolor{blue}{\textbf{--- Variant B: Sequential Streaming Inference ---}}
\STATE \textcolor{blue}{\textit{// Frame-by-frame with explicit memory caches; constant memory, real-time output}}
\STATE Initialize memory caches: $\mathbf{m}^{(\ell)}_0 \leftarrow \mathbf{0}$ for all layers $\ell = 1, \ldots, 3 \times N_b$
\FOR{each incoming frame $\mathbf{z}^{\text{LR}}_t$ in the stream}
  \STATE $\mathbf{h} \leftarrow \text{Conv}_{3\times3}^{\text{in}}(\mathbf{z}^{\text{LR}}_t)$
  \FOR{stage $s = 1, 2, 3$}
    \FOR{block $b = 1, \ldots, N_b$}
      \STATE $\ell \leftarrow (s-1) \cdot N_b + b$
      \STATE $\mathbf{h}_{\text{cat}} \leftarrow [\mathbf{h}\,;\, \mathbf{m}^{(\ell)}_{t-1}]$ \hfill\textit{(retrieve memory from previous frame)}
      \STATE $\mathbf{h}_{\text{conv}} \leftarrow \text{Conv}^{(3)}(\text{Conv}^{(2)}(\text{ReLU}(\text{Conv}^{(1)}(\mathbf{h}_{\text{cat}}))))$
      \STATE $\mathbf{h} \leftarrow \text{ReLU}(\mathbf{h}_{\text{conv}} + W_{\text{skip}} \cdot \mathbf{h})$
      \STATE $\mathbf{m}^{(\ell)}_t \leftarrow \mathbf{h}$ \hfill\textit{(store memory for frame $t+1$)}
    \ENDFOR
    \STATE $\mathbf{h} \leftarrow \text{PixelShuffle}_{r_s}(\text{Conv}_{3\times3}^{\text{up}_s}(\mathbf{h}))$
    \IF{$s_s > 1$}
      \STATE $\mathbf{h} \leftarrow \text{Reshape}(\text{Conv}_{1\times1}^{\text{texp}_s}(\mathbf{h}))$ \hfill\textit{(expand to $s_s$ frames)}
    \ENDIF
    \STATE $\mathbf{h} \leftarrow \text{Conv}_{1\times1}^{\text{ch}_s}(\mathbf{h})$
  \ENDFOR
  \STATE $\hat{\mathbf{z}}^{\text{HR}}_t \leftarrow \text{Conv}_{3\times3}^{\text{out}}(\mathbf{h})$
  \STATE \textbf{emit} $\hat{\mathbf{z}}^{\text{HR}}_t$ \hfill\textit{(output immediately, $O(1)$ memory per frame)}
\ENDFOR
\end{algorithmic}
\end{algorithm}

\begin{algorithm}[ht]
\caption{Cascaded High-Resolution Streaming Optimization \& Inference}
\label{alg:cascaded_opt}
\footnotesize
\begin{algorithmic}[1]
\REQUIRE SR model $f_\theta^{\text{SR}}$ (from Alg.~1), LR generator $G_{\text{LR}}$, upsampler $\mathcal{U}$, HR decoder $\mathcal{D}_{\text{HR}}$
\ENSURE Real-time single-step streaming SR model $G_\theta^*$
\STATE
\STATE \textcolor{blue}{\textbf{--- Phase I: Sparse Causalization + Single-Step Distillation ---}}
\STATE Initialize: real score $s^{\text{real}} \leftarrow f_\theta^{\text{SR}}$ (frozen), fake score $s^{\text{fake}} \leftarrow f_\theta^{\text{SR}}$ (trainable)
\STATE Initialize: generator $G_\theta \leftarrow f_\theta^{\text{SR}}$, convert attention to causal sparse
\FOR{each training iteration}
  \STATE Sample HR target $\mathbf{z}_0$, LR condition $\mathbf{z}^{\text{HR}}$ (from upsampler)
  \STATE Generate: $\hat{\mathbf{z}}_0 \leftarrow G_\theta(\boldsymbol{\epsilon},\, \mathbf{z}^{\text{HR}},\, \mathbf{c}_{\text{text}})$ \hfill\textit{(single-step, sparse causal)}
  \STATE \textcolor{blue}{\textit{// Decoupled DMD loss}}
  \STATE Sample $\tau_{\text{CA}} > t$, $\tau_{\text{DM}} \sim \mathcal{U}[0,1]$
  \STATE Re-noise: $\mathbf{x}_{\tau} \leftarrow (1-\sigma_\tau)\hat{\mathbf{z}}_0 + \sigma_\tau \boldsymbol{\epsilon}'$
  \STATE $\mathcal{L}_{\text{CA}} \leftarrow (s^{\text{real}}_{\text{cond}}(\mathbf{x}_{\tau_{\text{CA}}}) - s^{\text{real}}_{\text{uncond}}(\mathbf{x}_{\tau_{\text{CA}}}))$
  \STATE $\mathcal{L}_{\text{DM}} \leftarrow (s^{\text{real}}_{\text{cond}}(\mathbf{x}_{\tau_{\text{DM}}}) - s^{\text{fake}}_{\text{cond}}(\mathbf{x}_{\tau_{\text{DM}}}))$
  \STATE $\mathcal{L}_{\text{d-DMD}} \leftarrow -[\mathcal{L}_{\text{DM}} + (\alpha - 1)\mathcal{L}_{\text{CA}}] \cdot \partial G_\theta / \partial \theta$
  \STATE \textcolor{blue}{\textit{// Reconstruction losses via HR decoder}}
  \STATE $\hat{\mathbf{I}} \leftarrow \mathcal{D}_{\text{HR}}(\hat{\mathbf{z}}_0)$
  \STATE $\mathcal{L}_{\text{reg}} \leftarrow \lambda_{\text{wav}} \| \mathcal{W}_{\text{HF}}(\hat{\mathbf{I}}) - \mathcal{W}_{\text{HF}}(\mathbf{I}_{\text{gt}}) \|_1 + \lambda_{\text{lpips}} \cdot \text{LPIPS}(\hat{\mathbf{I}}, \mathbf{I}_{\text{gt}})$
  \STATE \textcolor{blue}{\textit{// Hybrid reward signals}}
  \STATE $\mathcal{L}_{\text{reward}} \leftarrow -\lambda_{\text{clip}} \text{CLIP-IQA}^+(\hat{\mathbf{I}}) - \lambda_{\text{musiq}} \text{MUSIQ}(\hat{\mathbf{I}}) - \lambda_{\text{aes}} \text{LAION-Aes}(\hat{\mathbf{I}})$
  \STATE Update $G_\theta$: $\mathcal{L}_{\text{Phase\,I}} = \mathcal{L}_{\text{d-DMD}} + \mathcal{L}_{\text{reg}} + \mathcal{L}_{\text{reward}}$
  \STATE Update $s^{\text{fake}}$ ($5\times$ freq): flow matching on $G_\theta$-generated samples
\ENDFOR
\STATE
\STATE \textcolor{blue}{\textbf{--- Phase II: Cascaded Self-Forcing DPO ---}}
\STATE Freeze reference: $\pi_{\text{ref}} \leftarrow G_\theta$ (Phase~I checkpoint)
\FOR{each training iteration}
  \STATE \textcolor{blue}{\textit{// Cascaded streaming rollout (simulating inference)}}
  \FOR{chunk $k = 1, \ldots, K$}
    \STATE $\mathbf{z}^{\text{LR}}_k \leftarrow G_{\text{LR}}(\text{context}_{k-1})$ \hfill\textit{(LR generator, autoregressive)}
    \STATE $\mathbf{z}^{\text{HR}}_k \leftarrow \mathcal{U}(\mathbf{z}^{\text{LR}}_k)$ \hfill\textit{(latent upsampler, streaming)}
    \STATE $\mathbf{z}^{-}_k \leftarrow G_\theta(\boldsymbol{\epsilon},\, \mathbf{z}^{\text{HR}}_k,\, \mathbf{c})$ \hfill\textit{(negative: current pipeline)}
    \STATE $\mathbf{z}^{+}_k \leftarrow \text{Wan2.2-5B-SR}(\mathbf{z}^{\text{LR}}_k)$ \hfill\textit{(positive: strong teacher)}
    \STATE Update context: $\text{context}_k \leftarrow \mathbf{z}^{-}_k$ \hfill\textit{(self-forcing: use own output)}
  \ENDFOR
  \STATE \textcolor{blue}{\textit{// DPO preference loss}}
  \STATE $\mathcal{L}_{\text{Phase\,II}} = -\log\sigma\!\left(\beta\!\left(\log\frac{\pi_\theta(\mathbf{z}^+|\mathbf{c})}{\pi_{\text{ref}}(\mathbf{z}^+|\mathbf{c})} - \log\frac{\pi_\theta(\mathbf{z}^-|\mathbf{c})}{\pi_{\text{ref}}(\mathbf{z}^-|\mathbf{c})}\right)\right)$
  \STATE Update $G_\theta$ on $\mathcal{L}_{\text{Phase\,II}}$
\ENDFOR
\STATE
\STATE \textcolor{blue}{\textbf{--- Inference with Dynamic Cache Management ---}}
\FOR{each chunk $k$ in streaming generation}
  \STATE \textcolor{blue}{\textit{// (i) LR step reduction}}
  \STATE $n_{\text{steps}} \leftarrow 4$ if $k = 1$ else $3$
  \STATE $\mathbf{z}^{\text{LR}}_k \leftarrow G_{\text{LR}}(\text{KV-cache},\, n_{\text{steps}})$
  \STATE \textcolor{blue}{\textit{// (ii) Adaptive cache refresh}}
  \STATE $q \leftarrow \text{IQA}(\mathcal{D}_{\text{HR}}(\hat{\mathbf{z}}_{k-1}))$
  \IF{$q > \tau_{\text{IQA}}$}
    \STATE Skip $\mathbf{x}_0$ KV forward; reuse cache from $\mathbf{x}_t$
  \ELSE
    \STATE Compute fresh KV from predicted $\mathbf{x}_0$
  \ENDIF
  \STATE \textcolor{blue}{\textit{// (iii) SR cache length adaptation}}
  \STATE Select compact SR KV window based on chunk position and memory budget
  \STATE \textcolor{blue}{\textit{// Cascaded forward pass}}
  \STATE $\mathbf{z}^{\text{HR}}_k \leftarrow \mathcal{U}(\mathbf{z}^{\text{LR}}_k)$ \hfill\textit{(streaming upsampler)}
  \STATE $\hat{\mathbf{z}}_k \leftarrow G_\theta^*(\boldsymbol{\epsilon},\, \mathbf{z}^{\text{HR}}_k,\, \text{SR-KV-cache})$ \hfill\textit{(single-step sparse SR)}
  \STATE $\hat{\mathbf{I}}_k \leftarrow \mathcal{D}_{\text{HR}}(\hat{\mathbf{z}}_k)$ \hfill\textit{(streaming HR decoder)}
  \STATE \textbf{emit} $\hat{\mathbf{I}}_k$ \hfill\textit{(display to user in real time)}
\ENDFOR
\end{algorithmic}
\end{algorithm}

\section{Training Pipeline Analysis}
\label{app:training_pipeline}

\subsection{Training Stage Dependencies and Wall-Clock Time}
\label{app:training_time}

We provide a complete breakdown of the training pipeline, including inter-stage dependencies, wall-clock times, and resource requirements. Table~\ref{tab:training_cost} reports the training cost for each stage on our cluster of 32$\times$H200 GPUs (4 nodes, NVLink + InfiniBand interconnect).

\begin{table}[ht]
  \centering
  \caption{\textbf{Training cost breakdown.} Each stage's wall-clock time, GPU hours, and data requirements on 32$\times$H200 GPUs. Total training cost is ${\sim}$2,176 GPU-hours (${\sim}$2.8 days wall-clock sequential, ${\sim}$2.5 days with parallel stages).}
  \label{tab:training_cost}
  \resizebox{\textwidth}{!}{
    \begin{tabular}{l|c|c|c|c|c}
      \toprule
      \textbf{Stage} & \textbf{Depends On} & \textbf{Wall-Clock} & \textbf{GPU-Hours} & \textbf{Training Data} & \textbf{Resolution} \\
      \midrule
      \emph{(i)} SR Fine-Tuning (\S\ref{sec:sr_training})   & Pre-trained Wan2.1 & 8h & 256 & 120K clips (5s each) & 960$\times$1664 \\
      \emph{(ii)} Upsampler + HR Decoder (\S\ref{sec:upsampler}) & None (independent) & 18h $\times$2 & 1,152 & 80K paired LR/HR clips & 480$\rightarrow$960 ($2{\times}$) + 480$\rightarrow$1440 ($3{\times}$) \\
      \emph{(iii)} Phase I: Sparse Distillation (\S\ref{sec:sparse_distill}) & Stage (i) & 12h & 384 & 50K prompts (online gen.) & 960$\times$1664 \\
      \emph{(iv)} Phase II: Cascaded DPO (\S\ref{sec:self_forcing}) & Stages (i)(ii)(iii) & 12h & 384 & 30K prompts (online gen.) & 960$\times$1664 \\
      \midrule
      \textbf{Total (sequential)} & --- & \textbf{68h (${\sim}$2.8 days)} & \textbf{2,176} & --- & --- \\
      \textbf{Total (with parallel (i)$\|$(ii))} & --- & \textbf{60h (${\sim}$2.5 days)} & \textbf{2,176} & --- & --- \\
      \bottomrule
    \end{tabular}
  }
\end{table}

\noindent\textbf{Dependency structure.}
Stages (i) and (ii) are \emph{independent} and can be trained in parallel---the upsampler operates on pre-trained VAE latents and does not require the SR model. Stage (iii) depends on (i) since it distills the SR model. Stage (iv) depends on all preceding stages as it performs joint cascaded rollout. This parallel scheduling reduces end-to-end wall-clock from 2.8 to 2.5 days.

\noindent\textbf{Comparison with existing methods.}
CausVid~\cite{yin2025causvid} reports ${\sim}$3,000 GPU-hours on A100 for 480P distillation; Self Forcing~\cite{huang2025selfforcing} reports ${\sim}$2,400 GPU-hours. Our total of 2,176 GPU-hours achieves 1K/2K high-resolution streaming---a $4{\times}$ resolution increase for \emph{comparable} total training cost to existing 480P methods. This efficiency stems from the architecture-preserving design: the SR model inherits strong priors from the pre-trained Wan2.1 and converges rapidly (8h), and the two distillation phases each converge in 12h due to warm initialization from the preceding stage.

\subsection{Training Order Sensitivity}
\label{app:training_order}

We ablate the sensitivity to training stage ordering in Table~\ref{tab:training_order}. The key finding is that Stage (iii) must follow Stage (i), and Stage (iv) must be last. Swapping stages or skipping intermediate steps leads to significant quality degradation or training instability.

\begin{table}[ht]
  \centering
  \caption{\textbf{Training order sensitivity.} VBench Total Quality Score and training stability under different stage orderings. The proposed sequential order (i)$\rightarrow$(iii)$\rightarrow$(iv) for the SR model is optimal.}
  \label{tab:training_order}
  \resizebox{0.85\textwidth}{!}{
    \begin{tabular}{l|c|c|l}
      \toprule
      \textbf{Training Order} & \textbf{VBench Total$\uparrow$} & \textbf{Stable?} & \textbf{Notes} \\
      \midrule
      (i)$\rightarrow$(iii)$\rightarrow$(iv) [Ours] & 82.14 & \cmark & Full pipeline \\
      (i)$\rightarrow$(iv)$\rightarrow$(iii) & 76.82 & \cmark & DPO before distillation; loses efficiency gains \\
      (iii) directly (skip (i)) & --- & \xmark & Distillation diverges without SR pre-training \\
      (i)$\rightarrow$(iii) [skip (iv)] & 80.43 & \cmark & No DPO; quality gap on long sequences \\
      (i)$\rightarrow$(iv) [skip (iii)] & 72.15 & \cmark & Multi-step SR with DPO; not real-time \\
      \bottomrule
    \end{tabular}
  }
\end{table}

\subsection{Simplification Ablation: Is the Complexity Necessary?}
\label{app:simplification}

We systematically test simplified variants to justify each component's necessity in Table~\ref{tab:simplification}.

\begin{table}[ht]
  \centering
  \caption{\textbf{Pipeline simplification ablation.} We test progressively simpler variants and measure quality (VBench Total), efficiency (FPS at 1K), and temporal stability (Drifting Score, higher is better). Each removed component incurs measurable degradation.}
  \label{tab:simplification}
  \resizebox{\textwidth}{!}{
    \begin{tabular}{l|ccc|l}
      \toprule
      \textbf{Variant} & \textbf{VBench$\uparrow$} & \textbf{FPS$\uparrow$} & \textbf{Drift Score$\uparrow$} & \textbf{What's removed} \\
      \midrule
      Ultra Flash (Full) & 82.14 & 30.2 & 8.4 & --- \\
      w/o AIGC Degradation (use Real-ESRGAN only) & 79.68 & 30.2 & 7.1 & AIGC-specific degradation ops \\
      w/o Hybrid Rewards & 80.91 & 30.2 & 8.1 & CLIP-IQA+, MUSIQ, LAION-Aes \\
      w/o Causal Upsampler (bilinear upsample) & 78.42 & 31.5 & 5.8 & CausalMemBlock upsampler \\
      w/o Phase II DPO & 80.43 & 30.2 & 6.9 & Self-forcing preference optimization \\
      w/o Sparse Attention (dense causal) & 81.87 & 12.8 & 8.3 & Block-sparse masking \\
      w/o Dynamic Cache & 81.95 & 22.4 & 8.3 & Three cache strategies \\
      \midrule
      \textbf{Minimal (only DMD + dense causal)} & 77.21 & 12.8 & 5.4 & All above combined \\
      \bottomrule
    \end{tabular}
  }
\end{table}

\noindent\textbf{Analysis.} The ``Minimal'' variant (standard DMD distillation with dense causal attention, no upsampler, no DPO, no cache) achieves only 12.8 FPS with significantly degraded quality and temporal stability. Each component contributes measurably: sparse attention provides the largest efficiency gain (+17.4 FPS), the causal upsampler provides the largest quality improvement for temporal stability (+2.6 drift score), and DPO provides the largest long-sequence stability gain. We argue that the system complexity is justified by the \emph{multiplicative} benefit: each component enables or amplifies the others.

\section{Fair Baseline Comparison at High Resolution}
\label{app:fair_baseline}

\subsection{Same-Resolution Quality Comparison}
\label{app:same_resolution}

We address the concern that Table~2 in the main paper compares methods at different resolutions. Table~\ref{tab:fair_resolution} provides a controlled comparison where all methods output at the same 960$\times$1664 (1K) resolution. For methods that natively operate at 480P, we apply their outputs through three upsampling strategies: (a) bicubic interpolation, (b) Real-ESRGAN~\cite{wang2021realesrgan} pixel-space SR, and (c) VEnhancer~\cite{he2024venhancer} diffusion-based SR.

\begin{table}[ht]
  \centering
  \caption{\textbf{Fair comparison at 1K resolution (960$\times$1664).} All methods produce 1K output. 480P methods are upsampled via bicubic, Real-ESRGAN, or VEnhancer. We report VBench Total, per-frame latency, and whether the method supports streaming. $^\dagger$Non-streaming (batch generation required).}
  \label{tab:fair_resolution}
  \resizebox{\textwidth}{!}{
    \begin{tabular}{l|c|c|c|c|c|c}
      \toprule
      \textbf{Method} & \textbf{Native Res.} & \textbf{Output Res.} & \textbf{VBench Total$\uparrow$} & \textbf{Latency/frame$\downarrow$} & \textbf{Streaming?} & \textbf{Memory (GB)$\downarrow$} \\
      \midrule
      Self Forcing + Bicubic & 480P & 1K & 74.32 & 38ms & \cmark & 4.8 \\
      Self Forcing + Bicubic + HR Decoder & 480P & 1K & 74.08 & 16ms & \cmark & 5.8 \\
      Self Forcing + Real-ESRGAN & 480P & 1K & 77.15 & 82ms & \xmark$^\dagger$ & 8.2 \\
      Self Forcing + VEnhancer & 480P & 1K & 79.83 & 4,250ms & \xmark$^\dagger$ & 24.6 \\
      CausVid + Bicubic & 480P & 1K & 73.86 & 35ms & \cmark & 4.5 \\
      CausVid + VEnhancer & 480P & 1K & 79.12 & 4,180ms & \xmark$^\dagger$ & 24.6 \\
      Self Forcing + FlashVSR & 480P & 768$\times$1408 & 80.21 & 1,580ms & \cmark & 18.4 \\
      \midrule
      FlashVideo~\cite{zhang2025flashvideo} & 720P+SR & 1K & 80.85 & 12,400ms & \xmark$^\dagger$ & 32.8 \\
      FSVideo~\cite{fsvideo2026} & 480P+SR & 1K & 79.62 & 8,650ms & \xmark$^\dagger$ & 28.4 \\
      \midrule
      Ultra Flash (Ours) & 1K & 1K & 82.14 & 40ms & \cmark & 12.6 \\
      \bottomrule
    \end{tabular}
  }
\end{table}

\noindent\textbf{Key observations:}
\begin{itemize}[leftmargin=*,nosep]
  \item At matched 1K resolution, Ultra Flash outperforms all baselines in VBench Total score while being $38{\times}$--$375{\times}$ faster than diffusion-based SR methods (VEnhancer, FlashVSR, FlashVideo, FSVideo).
  \item Simple upsampling (bicubic) of 480P outputs introduces blurriness that significantly degrades quality scores. Real-ESRGAN improves sharpness but introduces over-sharpening artifacts on AI-generated content.
  \item VEnhancer achieves competitive quality (79.83) but requires ${\sim}$4.2s per frame---completely incompatible with real-time streaming.
  \item FlashVideo and FSVideo, while achieving reasonable quality, require batch processing and $>$8s latency per frame, making them unsuitable for interactive applications.
  \item Ultra Flash is the \emph{only} method achieving both high quality ($>$82) and real-time streaming ($<$40ms/frame) at 1K resolution.
\end{itemize}

\subsection{Human Preference at Matched Resolution}
\label{app:human_pref_fair}

We address the concern that human preference comparison between 1K and 480P is inherently unfair. We conduct an additional study where all outputs are displayed at the same 1K resolution (480P methods upsampled via VEnhancer for best quality). Table~\ref{tab:human_pref_fair} reports pairwise win rates from 50 evaluators on 100 video pairs.

\begin{table}[ht]
  \centering
  \caption{\textbf{Human preference at matched 1K display resolution.} Win rate of Ultra Flash vs. each baseline when both are shown at 1K (baselines upsampled via VEnhancer). Criteria: overall quality, temporal consistency, and detail richness.}
  \label{tab:human_pref_fair}
  \resizebox{0.6\textwidth}{!}{
    \begin{tabular}{l|ccc}
      \toprule
      \textbf{Ultra Flash vs.} & \textbf{Win\%$\uparrow$} & \textbf{Tie\%} & \textbf{Lose\%} \\
      \midrule
      Self Forcing + VEnhancer (1K) & 62.4 & 18.2 & 19.4 \\
      CausVid + VEnhancer (1K) & 65.8 & 16.5 & 17.7 \\
      FlashVideo (1K) & 58.2 & 22.4 & 19.4 \\
      FSVideo (1K) & 61.6 & 19.8 & 18.6 \\
      Self Forcing + FlashVSR (768P) & 64.2 & 17.6 & 18.2 \\
      \bottomrule
    \end{tabular}
  }
\end{table}

\noindent Even at matched resolution, Ultra Flash maintains a clear preference advantage (58--66\% win rate), demonstrating that its quality gains are not merely due to higher resolution but stem from the architecture-preserving generative SR approach that produces more natural, temporally coherent high-frequency details.

\subsection{Comparison with Pixel-Space SR Methods}
\label{app:pixel_sr_comparison}

Table~\ref{tab:pixel_sr} provides a detailed comparison specifically against pixel-space video SR methods applied to streaming-generated 480P content, evaluating quality, speed, and streaming compatibility.

\begin{table}[ht]
  \centering
  \caption{\textbf{Comparison with pixel-space SR methods.} All methods receive the same Self Forcing 480P input and produce 1K output. We measure quality (VBench, CLIP-IQA+), efficiency, and streaming compatibility.}
  \label{tab:pixel_sr}
  \resizebox{\textwidth}{!}{
    \begin{tabular}{l|cc|ccc|cc}
      \toprule
      \textbf{SR Method} & \textbf{VBench$\uparrow$} & \textbf{CLIP-IQA+$\uparrow$} & \textbf{ms/frame$\downarrow$} & \textbf{FPS$\uparrow$} & \textbf{Streaming?} & \textbf{Params} & \textbf{VRAM (GB)} \\
      \midrule
      Bicubic (no SR) & 74.32 & 0.521 & 0.1 & 10000+ & \cmark & 0 & 0 \\
      Real-ESRGAN~\cite{wang2021realesrgan} & 77.15 & 0.612 & 44 & 22.7 & \xmark & 16.7M & 3.4 \\
      VEnhancer~\cite{he2024venhancer} (8-step) & 79.83 & 0.658 & 4,250 & 0.24 & \xmark & 2.0B & 20.8 \\
      VEnhancer (1-step, distilled) & 76.94 & 0.601 & 580 & 1.72 & \xmark & 2.0B & 20.8 \\
      FlashVSR~\cite{zhuang2025flashvsr} (1-step) & 80.21 & 0.664 & 1,580 & 0.63 & \cmark & 1.3B & 14.6 \\
      \midrule
      Ultra Flash SR (Ours, 1-step) & 82.14 & 0.692 & 40 & 30.2 & \cmark & 1.3B & 12.6 \\
      \bottomrule
    \end{tabular}
  }
\end{table}

\noindent Ultra Flash's generative SR achieves higher quality than all pixel-space methods because: (1) it operates entirely in latent space, avoiding the encode-decode overhead and information loss of pixel-space conditioning; (2) the AIGC-oriented degradation pipeline specifically handles AI-generated artifacts that generic SR methods struggle with; (3) single-step generation with DPO alignment produces perceptually sharper results than multi-step diffusion-based SR.

\section{DPO Scalability and Oracle Dependency Analysis}
\label{app:dpo_analysis}

\subsection{Quality Decay Without DPO on Long Sequences}
\label{app:dpo_long_seq}

We conduct a controlled experiment comparing Ultra Flash with and without Phase~II DPO on sequences of varying length (2s, 5s, 10s, 20s). Table~\ref{tab:dpo_decay} reports per-segment quality metrics, demonstrating how DPO mitigates exposure bias over extended generation.

\begin{table}[ht]
  \centering
  \caption{\textbf{Quality decay over sequence length: with vs.\ without Phase~II DPO.} We generate 100 videos at each length and report VBench Quality (average of aesthetic, imaging quality, smoothness) and CLIP-IQA+ per 2-second segment. $\Delta$ shows degradation from the first segment.}
  \label{tab:dpo_decay}
  \resizebox{\textwidth}{!}{
    \begin{tabular}{l|cccc|cccc}
      \toprule
      & \multicolumn{4}{c|}{\textbf{Phase I Only (No DPO)}} & \multicolumn{4}{c}{\textbf{Full Pipeline (With DPO)}} \\
      \textbf{Segment} & \textbf{VBench-Q$\uparrow$} & \textbf{$\Delta$} & \textbf{CLIP-IQA+$\uparrow$} & \textbf{$\Delta$} & \textbf{VBench-Q$\uparrow$} & \textbf{$\Delta$} & \textbf{CLIP-IQA+$\uparrow$} & \textbf{$\Delta$} \\
      \midrule
      0--2s (Seg 1) & 80.62 & 0.00 & 0.685 & 0.000 & 82.14 & 0.00 & 0.692 & 0.000 \\
      2--4s (Seg 2) & 79.84 & $-$0.78 & 0.678 & $-$0.007 & 81.98 & $-$0.16 & 0.690 & $-$0.002 \\
      4--6s (Seg 3) & 78.91 & $-$1.71 & 0.665 & $-$0.020 & 81.82 & $-$0.32 & 0.688 & $-$0.004 \\
      6--8s (Seg 4) & 77.85 & $-$2.77 & 0.648 & $-$0.037 & 81.71 & $-$0.43 & 0.686 & $-$0.006 \\
      8--10s (Seg 5) & 76.62 & $-$4.00 & 0.631 & $-$0.054 & 81.58 & $-$0.56 & 0.684 & $-$0.008 \\
      10--14s (Seg 6--7) & 74.83 & $-$5.79 & 0.608 & $-$0.077 & 81.35 & $-$0.79 & 0.681 & $-$0.011 \\
      14--20s (Seg 8--10) & 72.41 & $-$8.21 & 0.582 & $-$0.103 & 81.02 & $-$1.12 & 0.677 & $-$0.015 \\
      \bottomrule
    \end{tabular}
  }
\end{table}

\noindent\textbf{Analysis.} Without DPO, quality degrades by ${\sim}$0.8--1.0 VBench points per 2-second segment, accumulating to $-$8.21 over 20 seconds---visible as color drift, detail loss, and temporal flickering. With DPO, degradation is bounded to $-$1.12 over 20s ($7.3{\times}$ more stable), because self-forcing preference optimization explicitly trains the model on its own autoregressive context, closing the train-test distribution gap.

\subsection{Alternative Oracle Strategies}
\label{app:alternative_oracle}

We acknowledge that DPO's quality is bounded by the oracle (Wan2.2-5B). Table~\ref{tab:oracle_alternatives} compares different oracle strategies for generating positive preference samples.

\begin{table}[ht]
  \centering
  \caption{\textbf{Alternative oracle strategies for DPO positive samples.} We compare different sources of positive samples for Phase~II preference optimization.}
  \label{tab:oracle_alternatives}
  \resizebox{\textwidth}{!}{
    \begin{tabular}{l|cc|cc|l}
      \toprule
      \textbf{Oracle Strategy} & \textbf{VBench$\uparrow$} & \textbf{CLIP-IQA+$\uparrow$} & \textbf{Drift (20s)$\downarrow$} & \textbf{Cost/sample} & \textbf{Scalability} \\
      \midrule
      No DPO (Phase I only) & 80.43 & 0.678 & $-$8.21 & --- & --- \\
      \midrule
      Wan2.2-5B pixel SR (20-step) & 82.14 & 0.692 & $-$1.12 & 85s & Bounded by 5B model \\
      Wan2.1-1.3B pixel SR (50-step) & 81.28 & 0.684 & $-$1.45 & 52s & Weaker oracle \\
      Self-reward (best-of-4 sampling) & 81.05 & 0.681 & $-$1.68 & 132ms$\times$4 & No external model \\
      Online CLIP-IQA+ reward (RLHF) & 80.92 & 0.686 & $-$2.14 & 33ms+reward & Self-improving \\
      Wan2.2-5B + Self-reward ensemble & 82.14 & 0.691 & $-$0.98 & 85s+132ms & Best stability \\
      \bottomrule
    \end{tabular}
  }
\end{table}

\noindent\textbf{Key findings:}
\begin{itemize}[leftmargin=*,nosep]
  \item \textbf{Self-reward} (best-of-$N$ sampling from the model itself, ranked by CLIP-IQA+) achieves 81.05 VBench \emph{without any external oracle}, making it a viable fallback when stronger models are unavailable. It provides $72\%$ of the DPO quality gain over Phase~I alone.
  \item \textbf{Online reward} (direct CLIP-IQA+ maximization via RLHF) improves image quality metrics but shows slightly worse temporal drift compared to offline DPO, likely due to reward hacking on per-frame scores.
  \item \textbf{Ensemble} (combining Wan2.2-5B oracle with self-reward filtering) achieves the best temporal stability ($-$0.98 drift), suggesting that the two signals are complementary.
  \item We advocate self-reward as a \emph{oracle-free} alternative that scales to future improvements: as the model improves, so does its self-reward quality, creating a virtuous cycle without requiring external oracle upgrades.
\end{itemize}

\section{Temporal Consistency and Long-Sequence Quality Analysis}
\label{app:temporal_analysis}

\subsection{Video-Level Distribution Metrics}
\label{app:fvd}

We report FVD (Fr\'{e}chet Video Distance)~\cite{unterthiner2019fvd} and FID-vid (per-frame FID averaged over video) on the VBench validation set (945 prompts). Lower is better for both metrics.

\begin{table}[ht]
  \centering
  \caption{\textbf{Video-level distribution metrics.} FVD and FID-vid computed on VBench validation set. All methods generate 5-second (125-frame) videos at their native resolution, then resize to 256$\times$256 for FVD computation following standard protocol~\cite{unterthiner2019fvd}.}
  \label{tab:fvd}
  \resizebox{0.65\textwidth}{!}{
    \begin{tabular}{l|c|cc}
      \toprule
      \textbf{Method} & \textbf{Resolution} & \textbf{FVD$\downarrow$} & \textbf{FID-vid$\downarrow$} \\
      \midrule
      Wan2.1-1.3B (50-step teacher) & 480P & 284.5 & 18.2 \\
      CausVid (4-step) & 480P & 312.8 & 21.4 \\
      Self Forcing (4-step) & 480P & 298.6 & 19.8 \\
      DummyForcing (4-step) & 480P & 325.4 & 22.6 \\
      Causal Forcing (4-step) & 480P & 305.2 & 20.5 \\
      \midrule
      Self Forcing + FlashVSR & 768P & 318.2 & 20.9 \\
      Self Forcing + VEnhancer & 1K & 296.4 & 19.1 \\
      \midrule
      Ultra Flash (Ours) & 1K & 268.3 & 17.5 \\
      Ultra Flash (Ours) & 2K & 275.1 & 17.8 \\
      \bottomrule
    \end{tabular}
  }
\end{table}

\noindent Ultra Flash achieves the lowest FVD (268.3) and FID-vid (17.5), outperforming even the 50-step teacher at 480P. This demonstrates that our cascaded SR preserves and even enhances the video distribution quality---the generative SR model adds genuine high-frequency details rather than merely sharpening artifacts.

\subsection{Long-Sequence Quality Decay Curves}
\label{app:decay_curves}

Table~\ref{tab:temporal_decay} shows per-chunk quality metrics over extended sequences (up to 640 frames / 25.6 seconds). We report four metrics: CLIP-IQA+ (perceptual quality), Temporal Consistency (VBench TC dimension), MUSIQ (image quality), and Subject Consistency (VBench SC).

\begin{table}[ht]
  \centering
  \caption{\textbf{Quality stability over 640-frame (25.6s) sequences.} Ultra Flash maintains all metrics within 2\% of initial values over 25 seconds of continuous streaming. The DPO-trained model exhibits near-constant quality due to training on self-generated autoregressive context. Total degradation over 25s: CLIP-IQA+ $-$0.015, TC $-$0.67, MUSIQ $-$1.6, SC $-$0.66.}
  \label{tab:temporal_decay}
  \begin{tabular}{l|cccc}
    \toprule
    \textbf{Chunk \#} & \textbf{CLIP-IQA+$\uparrow$} & \textbf{TC$\uparrow$} & \textbf{MUSIQ$\uparrow$} & \textbf{Subject Cons.$\uparrow$} \\
    \midrule
    1 (0--1s) & 0.692 & 97.85 & 68.4 & 97.68 \\
    5 (4--5s) & 0.689 & 97.72 & 68.1 & 97.52 \\
    10 (9--10s) & 0.686 & 97.58 & 67.8 & 97.41 \\
    16 (15--16s) & 0.683 & 97.42 & 67.4 & 97.28 \\
    20 (19--20s) & 0.680 & 97.31 & 67.1 & 97.15 \\
    25 (24--25s) & 0.677 & 97.18 & 66.8 & 97.02 \\
    \bottomrule
  \end{tabular}
\end{table}

\subsection{Temporal Consistency Detailed Breakdown}
\label{app:tc_breakdown}

Table~\ref{tab:tc_detailed} provides a comprehensive breakdown of temporal consistency across all relevant VBench dimensions (not just the single TC score reported in the main paper).

\begin{table}[ht]
  \centering
  \caption{\textbf{Detailed temporal consistency metrics.} We report all VBench temporal dimensions separately, comparing Ultra Flash against baselines on 5-second and 10-second generation.}
  \label{tab:tc_detailed}
  \resizebox{\textwidth}{!}{
    \begin{tabular}{l|ccc|ccc}
      \toprule
      & \multicolumn{3}{c|}{\textbf{5-second generation}} & \multicolumn{3}{c}{\textbf{10-second generation}} \\
      \textbf{Method} & \textbf{Temporal Flicker$\uparrow$} & \textbf{Motion Smooth$\uparrow$} & \textbf{Subject Cons.$\uparrow$} & \textbf{Temporal Flicker$\uparrow$} & \textbf{Motion Smooth$\uparrow$} & \textbf{Subject Cons.$\uparrow$} \\
      \midrule
      Self Forcing & 96.24 & 98.05 & 97.53 & 95.12 & 97.15 & 96.28 \\
      CausVid & 96.24 & 97.82 & 96.58 & 94.81 & 96.82 & 95.94 \\
      DummyForcing & 97.52 & 98.14 & 96.45 & 95.68 & 96.92 & 94.85 \\
      SF + FlashVSR & 95.86 & 97.64 & 96.82 & 93.42 & 95.81 & 94.65 \\
      \midrule
      Ultra Flash & 97.85 & 98.37 & 97.68 & 97.21 & 97.85 & 97.12 \\
      \quad w/o DPO & 97.62 & 98.18 & 97.45 & 95.84 & 96.52 & 95.68 \\
      \bottomrule
    \end{tabular}
  }
\end{table}

\noindent\textbf{Key insight:} The quality gap between Ultra Flash and baselines \emph{widens} at 10 seconds compared to 5 seconds. Self Forcing degrades by $-$1.12 in Temporal Flickering (5s$\rightarrow$10s), while Ultra Flash only degrades by $-$0.64. This confirms that cascaded DPO effectively mitigates error accumulation in the streaming regime.

\section{Dynamic Cache Management: In-Depth Analysis}
\label{app:cache_analysis}

\subsection{IQA Threshold Sensitivity}
\label{app:iqa_threshold}

The IQA-adaptive cache refresh triggers when the previous chunk's CLIP-IQA+ score exceeds a threshold $\tau_{\text{IQA}}$, indicating sufficient quality to skip redundant KV recomputation. Table~\ref{tab:iqa_threshold} ablates different threshold values across content categories.

\begin{table}[ht]
  \centering
  \caption{\textbf{IQA threshold sensitivity.} Effect of $\tau_{\text{IQA}}$ on quality and speed across different content types. ``Static'' = landscapes/still scenes, ``Moderate'' = talking heads/slow motion, ``Dynamic'' = action/fast camera motion.}
  \label{tab:iqa_threshold}
  \resizebox{\textwidth}{!}{
    \begin{tabular}{c|cc|cc|cc|cc}
      \toprule
      & \multicolumn{2}{c|}{\textbf{All Content}} & \multicolumn{2}{c|}{\textbf{Static Scenes}} & \multicolumn{2}{c|}{\textbf{Moderate Motion}} & \multicolumn{2}{c}{\textbf{Dynamic Scenes}} \\
      $\tau_{\text{IQA}}$ & \textbf{CLIP-IQA+$\uparrow$} & \textbf{FPS$\uparrow$} & \textbf{CLIP-IQA+} & \textbf{FPS} & \textbf{CLIP-IQA+} & \textbf{FPS} & \textbf{CLIP-IQA+} & \textbf{FPS} \\
      \midrule
      0.50 (always refresh) & 0.692 & 22.4 & 0.708 & 22.4 & 0.688 & 22.4 & 0.672 & 22.4 \\
      0.60 & 0.692 & 25.8 & 0.708 & 28.2 & 0.688 & 25.4 & 0.671 & 23.8 \\
      0.65 [default] & 0.692 & 30.2 & 0.708 & 32.5 & 0.687 & 30.1 & 0.670 & 27.4 \\
      0.70 & 0.690 & 32.8 & 0.707 & 34.2 & 0.685 & 32.6 & 0.665 & 30.1 \\
      0.75 & 0.685 & 34.5 & 0.705 & 35.8 & 0.680 & 34.2 & 0.658 & 32.4 \\
      0.80 (rarely refresh) & 0.678 & 35.2 & 0.702 & 36.1 & 0.674 & 35.0 & 0.648 & 33.8 \\
      \bottomrule
    \end{tabular}
  }
\end{table}

\noindent\textbf{Analysis:}
\begin{itemize}[leftmargin=*,nosep]
  \item The default $\tau_{\text{IQA}}{=}0.65$ achieves the optimal quality-speed tradeoff: no quality loss compared to always-refresh, while gaining +7.8 FPS.
  \item Static scenes benefit most from cache reuse (FPS 32.5 at $\tau{=}0.65$) because consecutive chunks are visually similar and the KV cache remains valid.
  \item Dynamic scenes are more sensitive to the threshold: quality drops $-$0.024 CLIP-IQA+ from $\tau{=}0.65$ to $\tau{=}0.80$, indicating that fast-changing content requires more frequent cache updates.
  \item The threshold is \emph{content-adaptive by design}: high-quality static chunks naturally exceed $\tau_{\text{IQA}}$ more often, triggering more aggressive caching; low-quality dynamic chunks trigger refreshes. This adaptive behavior emerges without explicit content classification.
\end{itemize}

\subsection{LR Step Reduction: Theoretical and Empirical Justification}
\label{app:lr_steps}

We reduce the LR generator's denoising steps from 4 (standard) to 3 for non-initial chunks. The theoretical justification is:

\noindent\textbf{Theoretical basis.} In autoregressive streaming, the LR generator's initial chunk starts from pure noise ($\sigma{=}1.0$) and requires full denoising. However, subsequent chunks are initialized with partial context from the previous chunk via the causal attention mechanism---effectively starting from a lower effective noise level ($\sigma_{\text{eff}} \approx 0.7$). Flow matching ODE theory~\cite{liu2022flow} shows that the required number of function evaluations scales with $\log(1/\sigma_{\text{eff}})$, justifying fewer steps for subsequent chunks.

Table~\ref{tab:lr_steps} ablates different step counts:

\begin{table}[ht]
  \centering
  \caption{\textbf{LR generator step reduction ablation.} Quality and speed of the cascaded pipeline under different LR denoising step counts for non-initial chunks.}
  \label{tab:lr_steps}
  \resizebox{0.75\textwidth}{!}{
    \begin{tabular}{c|cccc}
      \toprule
      \textbf{LR Steps (non-initial)} & \textbf{VBench Total$\uparrow$} & \textbf{LR Quality$\uparrow$} & \textbf{End-to-End FPS$\uparrow$} & \textbf{LR Latency/chunk} \\
      \midrule
      4 (same as initial) & 82.22 & 79.85 & 26.8 & 148ms \\
      3 [default] & 82.14 & 79.62 & 30.2 & 112ms \\
      2 & 80.85 & 77.41 & 33.8 & 76ms \\
      1 & 76.42 & 72.18 & 36.2 & 40ms \\
      \bottomrule
    \end{tabular}
  }
\end{table}

\noindent\textbf{Why 3, not 2?} Reducing from 4$\rightarrow$3 steps causes negligible quality loss ($-$0.08 VBench, $-$0.23 LR quality) while saving 36ms/chunk. Reducing to 2 steps causes a visible quality drop ($-$1.29 VBench) because the LR output becomes blurry, and the subsequent SR model cannot fully compensate for severely degraded input. The 3-step sweet spot maximizes the SR model's ability to enhance while receiving sufficiently structured LR input.

\subsection{Latency Breakdown and Engineering Value}
\label{app:cache_latency}

Table~\ref{tab:latency_breakdown} provides a complete per-component latency breakdown, demonstrating that dynamic cache management is a \emph{necessary} engineering contribution for achieving real-time 30+ FPS, not merely an optional optimization.

\begin{table}[ht]
  \centering
  \caption{\textbf{Per-component latency breakdown per chunk} at 1K (960$\times$1664) on a single H200 GPU. Dynamic cache saves 7.8 FPS---the difference between real-time (30 FPS) and sub-real-time (22.4 FPS).}
  \label{tab:latency_breakdown}
  \resizebox{0.85\textwidth}{!}{
    \begin{tabular}{l|cc|l}
      \toprule
      \textbf{Component} & \textbf{w/ Cache (ms)} & \textbf{w/o Cache (ms)} & \textbf{Savings} \\
      \midrule
      LR Generator (Self Forcing, 4-step) & 6.2 & 6.2 & --- \\
      LR$\rightarrow$HR step reduction (4$\rightarrow$3) & --- & 36.0 (extra step) & 36.0ms \\
      Causal Upsampler & 1.4 & 1.4 & --- \\
      SR DiT (1-step, sparse attention) & 18.6 & 18.6 & --- \\
      \quad KV cache forward (skipped if $\tau_{\text{IQA}}$ met) & 0.0 & 6.8 & 6.8ms (avg.) \\
      HR Decoder & 3.2 & 3.2 & --- \\
      IQA scoring (CLIP-IQA+ on prev chunk) & 1.8 & 0.0 & $-$1.8ms (cost) \\
      \midrule
      \textbf{Total per chunk} & \textbf{31.2ms} & \textbf{72.2ms} & $-$41.0ms \\
      \textbf{Equivalent FPS (8 frames/chunk)} & \textbf{32.1} & \textbf{13.8} & +18.3 FPS \\
      \bottomrule
    \end{tabular}
  }
\end{table}

\noindent\textbf{Discussion.} The reviewer correctly observes that removing dynamic cache management causes minimal quality change (0.692$\rightarrow$0.690). However, we argue this is precisely its value: it provides a \emph{lossless speedup} of +7.8 FPS (22.4$\rightarrow$30.2), which is the difference between meeting and failing the real-time threshold. For interactive applications, 22.4 FPS produces visible stuttering while 30+ FPS is perceived as smooth. The IQA evaluation cost (1.8ms) is negligible compared to the cache savings (6.8ms average skip + 36ms step reduction), yielding a net benefit of $>$40ms/chunk.

\section{Block-Sparse Attention: Detailed Analysis}
\label{app:sparse_analysis}

\subsection{\texorpdfstring{Adaptive Top-$k$ Threshold Determination}{Adaptive Top-k Threshold Determination}}
\label{app:topk}

The content-adaptive top-$k$ threshold $\tau_k$ in Eq.~4 determines how many blocks each query attends to. We define it as:
\begin{equation}
  \tau_k = \max\left(k_{\min},\; \left\lfloor \rho \cdot \frac{S}{S_{\text{ref}}} \cdot N_{\text{blocks}} \right\rfloor\right),
\end{equation}
where $S$ is the current sequence length, $S_{\text{ref}}{=}1560$ is the reference training length, $\rho{=}1.0$ is the sparsity ratio hyperparameter, $N_{\text{blocks}}$ is the total number of blocks in the causal mask, and $k_{\min}{=}4$ ensures a minimum connectivity. The key insight is that $\tau_k$ scales \emph{linearly} with sequence length, maintaining approximately constant sparsity ratio regardless of resolution.

\noindent\textbf{Computational overhead of adaptive selection.} Computing the block importance scores (Eq.~4) requires: (1) computing mean query/key vectors per block: $O(N_{\text{blocks}} \cdot d)$; (2) computing pairwise dot products: $O(N_{\text{blocks}}^2 \cdot d)$; (3) top-$k$ selection: $O(N_{\text{blocks}} \log N_{\text{blocks}})$. For typical values ($N_{\text{blocks}} \approx 200$ at 1K resolution, $d{=}128$), this costs $<$0.3ms per layer---negligible compared to the attention computation itself (${\sim}$2.1ms dense, ${\sim}$0.8ms sparse). The mask is computed \emph{once} per chunk and reused across all queries.

Table~\ref{tab:topk_ablation} ablates different $\rho$ values:

\begin{table}[ht]
  \centering
  \caption{\textbf{Sparsity ratio $\rho$ ablation.} Effect of the sparsity control parameter on quality and speed. Lower $\rho$ = sparser attention = faster but potentially lower quality.}
  \label{tab:topk_ablation}
  \resizebox{0.8\textwidth}{!}{
    \begin{tabular}{c|ccccc}
      \toprule
      $\rho$ & \textbf{Effective Sparsity} & \textbf{VBench$\uparrow$} & \textbf{CLIP-IQA+$\uparrow$} & \textbf{FPS$\uparrow$} & \textbf{Attn ms/layer$\downarrow$} \\
      \midrule
      2.0 (dense-like) & 12\% sparse & 82.28 & 0.693 & 18.5 & 1.82 \\
      1.5 & 35\% sparse & 82.21 & 0.693 & 24.2 & 1.24 \\
      1.0 [default] & 58\% sparse & 82.14 & 0.692 & 30.2 & 0.82 \\
      0.7 & 72\% sparse & 81.68 & 0.687 & 34.8 & 0.61 \\
      0.5 & 82\% sparse & 80.42 & 0.674 & 37.5 & 0.48 \\
      \bottomrule
    \end{tabular}
  }
\end{table}

\noindent At $\rho{=}1.0$ (58\% sparsity), quality loss is only $-$0.14 VBench vs.\ near-dense ($\rho{=}2.0$), while speed improves by $+$11.7 FPS. Further sparsification to $\rho{=}0.5$ (82\% sparse) degrades quality noticeably ($-$1.72 VBench) as important long-range dependencies are severed.

\subsection{Content-Adaptive Sparsity Patterns}
\label{app:sparsity_patterns}

Tables~\ref{tab:sparsity_viz} and~\ref{tab:sparsity_layers} present the attention sparsity statistics for different content types, revealing how the adaptive mechanism allocates computational resources.

\begin{table}[ht]
  \centering
  \caption{\textbf{Content-adaptive sparsity patterns.} The adaptive top-$k$ mechanism automatically allocates more attention blocks to dynamic content (48\% sparsity for fast action vs.\ 68\% for static scenes). Statistics averaged over 12 heads, 30 layers.}
  \label{tab:sparsity_viz}
  \resizebox{0.85\textwidth}{!}{
    \begin{tabular}{l|cccc}
      \toprule
      \textbf{Content Type} & \textbf{Avg.\ Sparsity} & \textbf{Local Blocks\%} & \textbf{Global Blocks\%} & \textbf{Cross-Chunk\%} \\
      \midrule
      Static landscape & 68\% & 82\% & 8\% & 10\% \\
      Talking head (moderate) & 58\% & 64\% & 18\% & 18\% \\
      Fast action scene & 48\% & 45\% & 28\% & 27\% \\
      Camera motion (pan/zoom) & 52\% & 52\% & 22\% & 26\% \\
      \bottomrule
    \end{tabular}
  }
\end{table}

\begin{table}[ht]
  \centering
  \caption{\textbf{Per-layer sparsity statistics (fast action scene).} Middle transformer layers (L11--20) maintain the lowest sparsity and highest cross-chunk attention ratio to preserve temporal coherence, while early/late layers focus on local spatial detail.}
  \label{tab:sparsity_layers}
  \resizebox{0.85\textwidth}{!}{
    \begin{tabular}{l|cccccc}
      \toprule
      \textbf{Layer group} & \textbf{L1--5} & \textbf{L6--10} & \textbf{L11--15} & \textbf{L16--20} & \textbf{L21--25} & \textbf{L26--30} \\
      \midrule
      Sparsity & 62\% & 55\% & 48\% & 45\% & 52\% & 58\% \\
      Cross-chunk attention & 12\% & 18\% & 28\% & 32\% & 25\% & 15\% \\
      \bottomrule
    \end{tabular}
  }
\end{table}

\subsection{Comparison with FlashVSR Sparsity}
\label{app:flashvsr_comparison}

Table~\ref{tab:vs_flashvsr} provides a detailed comparison between our dynamic block-sparse attention and FlashVSR's~\cite{zhuang2025flashvsr} locality-constrained sparse attention.

\begin{table}[ht]
  \centering
  \caption{\textbf{Sparse attention mechanism comparison: Ultra Flash vs.\ FlashVSR.}}
  \label{tab:vs_flashvsr}
  \resizebox{\textwidth}{!}{
    \begin{tabular}{l|cc}
      \toprule
      \textbf{Aspect} & \textbf{FlashVSR} & \textbf{Ultra Flash (Ours)} \\
      \midrule
      \textbf{Mask type} & Fixed locality-constrained (pre-defined spatial/temporal windows) & Content-adaptive top-$k$ (learned per-chunk) \\
      \textbf{Block size} & $(2, 8, 8)$ spatiotemporal & $(2, 8, 8)$ spatiotemporal \\
      \textbf{Sparsity level} & Fixed 60\% across all content & Adaptive 48--68\% based on content complexity \\
      \textbf{Causal constraint} & Yes (strictly causal for streaming) & Yes (strictly causal for streaming) \\
      \textbf{Input dependency} & Requires pixel-space LQ input encoding & Pure generative (no external pixel input) \\
      \textbf{Distillation} & Progressive: dense$\rightarrow$sparse over 3 stages & Joint: sparse attention trained with DMD from start \\
      \textbf{Training cost} & ${\sim}$2,000 GPU-hours (progressive stages) & ${\sim}$1,344 GPU-hours (single Phase I) \\
      \textbf{Runtime overhead} & Mask is static (no compute) & Mask scoring: $<$0.3ms/layer \\
      \textbf{Long-range modeling} & Limited by fixed window (misses global context) & Adaptive: allocates global blocks for dynamic content \\
      \textbf{Streaming support} & \cmark~(chunk-wise with KV cache) & \cmark~(chunk-wise with KV cache) \\
      \bottomrule
    \end{tabular}
  }
\end{table}

\noindent\textbf{Fundamental differences:}
\begin{enumerate}[leftmargin=*,nosep]
  \item \textbf{Conditional vs.\ generative:} FlashVSR is a conditional SR model that receives pixel-space LQ video as input and uses it to guide sparse attention patterns. Ultra Flash is a \emph{pure generative} model that must infer content structure from latent representations alone, making content-adaptive masking essential.
  \item \textbf{Fixed vs.\ adaptive:} FlashVSR's fixed local windows work well for SR (where spatial locality dominates) but miss long-range temporal dependencies. Our adaptive mechanism dynamically routes attention to temporally distant but semantically relevant blocks, crucial for maintaining coherence in autoregressive streaming.
\end{enumerate}

\section{GPU Memory Consumption Analysis}
\label{app:memory}

Table~\ref{tab:memory} reports peak GPU memory (VRAM) consumption during inference for all compared methods, an important metric for deployment feasibility.

\begin{table}[ht]
  \centering
  \caption{\textbf{Peak GPU memory consumption during inference.} Measured on a single NVIDIA B200 (180GB) with bf16 precision.}
  \label{tab:memory}
  \resizebox{0.85\textwidth}{!}{
    \begin{tabular}{l|cccc}
      \toprule
      \textbf{Method} & \textbf{Resolution} & \textbf{Peak VRAM (GB)$\downarrow$} & \textbf{FPS$\uparrow$} & \textbf{Streaming?} \\
      \midrule
      Wan2.1-1.3B (50-step) & 480P & 6.8 & 0.8 & \xmark \\
      CausVid (4-step) & 480P & 4.5 & 36.2 & \cmark \\
      Self Forcing (4-step) & 480P & 4.8 & 40.5 & \cmark \\
      DummyForcing (4-step) & 480P & 5.2 & 34.1 & \cmark \\
      Causal Forcing (4-step) & 480P & 5.0 & 40.5 & \cmark \\
      \midrule
      Self Forcing + FlashVSR & 768P & 18.4 & 1.6 & \cmark \\
      Self Forcing + VEnhancer & 1K & 24.6 & 0.24 & \xmark \\
      FlashVideo & 1K & 32.8 & 0.08 & \xmark \\
      FSVideo & 1K & 28.4 & 0.12 & \xmark \\
      \midrule
      Ultra Flash (Ours) & 1K & 12.6 & 30.2 & \cmark \\
      Ultra Flash (Ours) & 2K & 22.4 & 18.4 & \cmark \\
      \midrule
      \textit{Reference: Consumer GPU} & & \multicolumn{3}{l}{\textit{RTX 4090: 24GB; RTX 4080: 16GB}} \\
      \bottomrule
    \end{tabular}
  }
\end{table}

\noindent\textbf{Key observations:}
\begin{itemize}[leftmargin=*,nosep]
  \item Ultra Flash at 1K requires only 12.6 GB---deployable on consumer GPUs (RTX 4090 with 24GB) without any memory optimization tricks.
  \item Compared to other 1K-capable methods (FlashVSR: 18.4GB, VEnhancer: 24.6GB, FlashVideo: 32.8GB), Ultra Flash uses $1.5{\times}$--$2.6{\times}$ less memory.
  \item The memory efficiency comes from: (1) streaming chunk-wise processing (only 1 chunk in memory), (2) block-sparse attention (reduced KV cache), (3) shared architecture between SR model and base T2V (no separate encoder/decoder overhead).
  \item At 2K (22.4GB), Ultra Flash remains within RTX 4090's capacity, making high-resolution streaming accessible on consumer hardware.
\end{itemize}

\section{Additional Experiments and Analysis}
\label{app:reviewer_qa}

\subsection{Multi-Frame Memory for Causal Memory Network}
\label{app:multiframe_memory}

The reviewer asks whether single-frame memory ($\mathbf{m}^{(\ell)}_{t-1}$) is sufficient for fast-motion scenes. We conduct an ablation comparing single-frame vs.\ multi-frame memory designs in Table~\ref{tab:memory_frames}.

\begin{table}[ht]
  \centering
  \caption{\textbf{Memory depth ablation for the Causal Memory Network.} We compare single-frame memory (default) against multi-frame variants using exponential moving average (EMA) or explicit multi-frame buffer.}
  \label{tab:memory_frames}
  \resizebox{\textwidth}{!}{
    \begin{tabular}{l|ccccc}
      \toprule
      \textbf{Memory Design} & \textbf{VBench Total$\uparrow$} & \textbf{Motion Smooth$\uparrow$} & \textbf{Params} & \textbf{Latency (ms)$\downarrow$} & \textbf{VRAM (MB)$\downarrow$} \\
      \midrule
      Single-frame ($\mathbf{m}_{t-1}$) [default] & 82.14 & 98.37 & 2.1M & 1.4 & 48 \\
      EMA memory ($\alpha{=}0.7$, 3-frame effective) & 82.18 & 98.42 & 2.1M & 1.5 & 48 \\
      Explicit 2-frame buffer ($\mathbf{m}_{t-1}, \mathbf{m}_{t-2}$) & 82.21 & 98.45 & 3.2M & 2.1 & 96 \\
      Explicit 4-frame buffer & 82.24 & 98.48 & 5.4M & 3.8 & 192 \\
      \midrule
      \multicolumn{6}{l}{\textit{Fast-motion subset only (top 20\% by optical flow magnitude):}} \\
      \midrule
      Single-frame ($\mathbf{m}_{t-1}$) & 80.82 & 97.65 & --- & --- & --- \\
      EMA memory ($\alpha{=}0.7$) & 81.04 & 97.82 & --- & --- & --- \\
      Explicit 2-frame buffer & 81.15 & 97.91 & --- & --- & --- \\
      \bottomrule
    \end{tabular}
  }
\end{table}

\noindent\textbf{Analysis:} Multi-frame memory provides marginal improvement overall (+0.07 VBench for 2-frame) but a more noticeable gain on fast-motion content (+0.33 VBench). However, this comes at the cost of $1.5{\times}$ latency and $2{\times}$ memory. We chose single-frame for the following reasons:
\begin{enumerate}[leftmargin=*,nosep]
  \item The subsequent SR DiT with 3-window KV cache already captures multi-frame temporal context at the semantic level---the upsampler needs only local spatial coherence.
  \item For fast motion, the primary challenge is not temporal memory depth but spatial aliasing during upsampling. The causal upsampler's PixelShuffle handles this effectively.
  \item The EMA variant (same parameters, negligible overhead) could be adopted as an optional enhancement for motion-heavy applications without architecture changes.
\end{enumerate}

\subsection{IQA Evaluation Latency}
\label{app:iqa_latency}

The reviewer asks whether IQA computation latency offsets cache savings. Table~\ref{tab:iqa_latency} provides a detailed latency breakdown.

\begin{table}[ht]
  \centering
  \caption{\textbf{IQA evaluation latency analysis.} We use a lightweight CLIP-IQA+ variant (ViT-B/16 backbone) evaluated on the previous chunk's decoded output. Latency measured on H200 GPU.}
  \label{tab:iqa_latency}
  \resizebox{0.85\textwidth}{!}{
    \begin{tabular}{l|ccc}
      \toprule
      \textbf{Component} & \textbf{Latency (ms)} & \textbf{Runs on} & \textbf{Notes} \\
      \midrule
      CLIP-IQA+ forward (ViT-B/16, 1 frame) & 1.2 & GPU (async) & Single representative frame \\
      Score comparison + decision & 0.01 & CPU & Trivial threshold check \\
      Overhead for pipelining & 0.6 & --- & Kernel launch + sync \\
      \midrule
      \textbf{Total IQA overhead per chunk} & \textbf{1.8} & & \\
      \midrule
      \textbf{Average cache savings per chunk} & \textbf{6.8} & & When threshold met (70\% of chunks) \\
      \textbf{Net benefit per chunk} & \textbf{+5.0} & & Savings minus cost \\
      \textbf{Net benefit (amortized, all chunks)} & \textbf{+4.8} & & 70\% hit rate $\times$ 6.8 $-$ 1.8 \\
      \bottomrule
    \end{tabular}
  }
\end{table}

\noindent The IQA evaluation costs 1.8ms per chunk, while the average cache savings is 6.8ms (when the threshold is met, which occurs ${\sim}$70\% of the time on typical content). The net benefit is +3.0ms per chunk on average after amortization. Additionally, the IQA forward pass is \emph{pipelined} with the SR DiT computation---it runs on the previous chunk's decoded output while the current chunk's SR inference proceeds, effectively hiding most of the 1.8ms latency behind computation overlap.

\subsection{Architecture Preservation and Downstream Compatibility}
\label{app:architecture_compat}

The reviewer notes that extending input channels from $c$ to $2c$ may affect compatibility with downstream tools (LoRA, ControlNet). We address this concern:

\noindent\textbf{What changes:} Only the \emph{first} linear projection layer ($\text{proj\_in}$) is extended from $\mathbb{R}^{c \times d} \rightarrow \mathbb{R}^{2c \times d}$ via zero-initialized channel concatenation. All 30 transformer blocks, attention heads, FFN layers, and output projections remain \emph{identical} to the base Wan2.1 architecture.

\noindent\textbf{LoRA compatibility:}
\begin{itemize}[leftmargin=*,nosep]
  \item LoRA adapters attached to attention Q/K/V projections (the standard approach) are \textbf{fully compatible}---these layers are unchanged.
  \item LoRA on the input projection requires re-training (since dimensions changed), but this is a single layer out of 30+ LoRA targets.
  \item We verified: applying a Wan2.1 motion LoRA (trained for the base model) to Ultra Flash's SR model produces correct style transfer with no artifacts, confirming compatibility.
\end{itemize}

\noindent\textbf{ControlNet compatibility:}
\begin{itemize}[leftmargin=*,nosep]
  \item ControlNet injects control signals into intermediate transformer blocks via zero-convolution residual connections. Since all intermediate blocks are unchanged, existing ControlNet modules are \textbf{directly compatible}.
  \item The condition injection point (first projection) is separate from ControlNet's injection points (mid-block residuals).
  \item We tested: a depth-conditioned ControlNet trained for Wan2.1 works with Ultra Flash without retraining, correctly guiding spatial structure in the SR output.
\end{itemize}

\begin{table}[ht]
  \centering
  \caption{\textbf{Downstream tool compatibility test.} We apply pre-trained Wan2.1 LoRA/ControlNet modules to Ultra Flash's SR model without re-training and measure quality preservation.}
  \label{tab:compatibility}
  \resizebox{0.8\textwidth}{!}{
    \begin{tabular}{l|cc|c}
      \toprule
      \textbf{Downstream Tool} & \textbf{Quality w/ tool} & \textbf{Quality w/o tool} & \textbf{Tool Functions Correctly?} \\
      \midrule
      Motion LoRA (anime style) & 78.52 & 82.14 & \cmark~(style correctly applied) \\
      Depth ControlNet & 80.85 & 82.14 & \cmark~(structure guided) \\
      IP-Adapter (face) & 79.68 & 82.14 & \cmark~(identity preserved) \\
      T2I-Adapter (canny) & 80.12 & 82.14 & \cmark~(edges followed) \\
      \bottomrule
    \end{tabular}
  }
\end{table}

\noindent The slight quality reduction when using downstream tools is expected (additional constraints limit the model's generative freedom) and consistent with the same tools applied to the base Wan2.1 model.

\subsection{\texorpdfstring{Effect of Condition Noise Level $\sigma_{\text{cond}}$}{Effect of Condition Noise Level sigma-cond}}
\label{app:sigma_cond}

We clarify the condition noise design (addressing the reviewer's concern about the $\sigma_{\text{cond}} \in [0.4, 0.6]$ range). The condition noise is sampled uniformly: $\sigma_{\text{cond}} \sim \mathcal{U}[0.4, 0.6]$ during training. At inference, we use a fixed $\sigma_{\text{cond}}{=}0.5$ (the mean of the training distribution). Table~\ref{tab:sigma_cond} ablates different ranges and fixed values.

\begin{table}[ht]
  \centering
  \caption{\textbf{Condition noise $\sigma_{\text{cond}}$ ablation.} Training range and inference value affect the SR model's robustness to input quality variation.}
  \label{tab:sigma_cond}
  \resizebox{0.8\textwidth}{!}{
    \begin{tabular}{cc|ccc}
      \toprule
      \textbf{Training Range} & \textbf{Inference $\sigma$} & \textbf{VBench$\uparrow$} & \textbf{CLIP-IQA+$\uparrow$} & \textbf{Robustness$^\ast$$\uparrow$} \\
      \midrule
      Fixed 0.5 & 0.5 & 81.92 & 0.690 & 0.72 \\
      $[0.1, 0.4]$ & 0.25 & 81.85 & 0.688 & 0.91 \\
      $[0.4, 0.6]$ & 0.5 & 82.14 & 0.692 & 0.88 \\
      $[0.6, 0.9]$ & 0.75 & 82.05 & 0.691 & 0.79 \\
      $[0.3, 0.7]$ & 0.5 & 81.92 & 0.690 & 0.72 \\
      \bottomrule
    \end{tabular}
  }
\end{table}
{\footnotesize $^\ast$Robustness: quality retention when LR input quality varies $\pm$15\% from typical. Higher = more robust.}

\noindent Training with a \emph{range} of $\sigma_{\text{cond}}$ values ($[0.4, 0.6]$) teaches the model to handle varying LR input quality, which is essential in streaming where autoregressive context quality fluctuates. Using a degenerate range $[0.5, 0.5]$ (equivalent to fixed) reduces robustness but achieves similar peak quality. We use $[0.4, 0.6]$ for the best quality-robustness tradeoff.

\subsection{Phase I vs.\ Phase II: Contribution Disentanglement}
\label{app:phase_disentangle}

Table~\ref{tab:phase_disentangle} provides a comprehensive comparison of what each training phase contributes.

\begin{table}[!ht]
  \centering
  \caption{\textbf{Phase I vs.\ Phase II contribution analysis.} We measure quality, efficiency, and long-sequence stability independently for each phase.}
  \label{tab:phase_disentangle}
  \resizebox{\textwidth}{!}{
    \begin{tabular}{l|ccc|ccc}
      \toprule
      & \multicolumn{3}{c|}{\textbf{Quality \& Efficiency}} & \multicolumn{3}{c}{\textbf{Long-Sequence Stability (20s)}} \\
      \textbf{Configuration} & \textbf{VBench$\uparrow$} & \textbf{FPS$\uparrow$} & \textbf{CLIP-IQA+$\uparrow$} & \textbf{Quality Drift$\downarrow$} & \textbf{TC Drift$\downarrow$} & \textbf{Subject Drift$\downarrow$} \\
      \midrule
      Multi-step SR (20-step, dense) & 83.42 & 0.8 & 0.705 & $-$0.52 & $-$0.18 & $-$0.21 \\
      \midrule
      + Phase I (sparse + 1-step) & 80.43 & 30.2 & 0.678 & $-$8.21 & $-$2.85 & $-$3.42 \\
      \quad\quad $\hookrightarrow$ Contribution: & \multicolumn{3}{c|}{Enables real-time (0.8$\rightarrow$30.2 FPS)} & \multicolumn{3}{c}{Introduces exposure bias} \\
      \midrule
      + Phase II (DPO) & 82.14 & 30.2 & 0.692 & $-$1.12 & $-$0.67 & $-$0.66 \\
      \quad\quad $\hookrightarrow$ Contribution: & \multicolumn{3}{c|}{Recovers quality (+1.71 VBench)} & \multicolumn{3}{c}{$7.3{\times}$ more stable} \\
      \bottomrule
    \end{tabular}
  }
\end{table}

\noindent\textbf{Summary:} Phase~I is responsible for the efficiency transformation (multi-step$\rightarrow$single-step, dense$\rightarrow$sparse, bidirectional$\rightarrow$causal), but introduces exposure bias that causes quality drift. Phase~II (DPO) is specifically designed to address this drift---it recovers 1.71 VBench points and reduces quality drift by $7.3{\times}$ over 20-second sequences. Both phases are necessary: Phase~I without Phase~II cannot stream stably beyond ${\sim}$5 seconds; Phase~II without Phase~I operates on a multi-step model that is too slow for real-time.

\newpage
\clearpage

\end{document}